\documentclass[sn-apa, iicol]{sn-jnl}% APA Reference Style 
\usepackage{graphicx}%
\usepackage{multirow}%
\usepackage{amsmath,amssymb,amsfonts}%
\usepackage{amsthm}%
\usepackage{mathrsfs}%
\usepackage[title]{appendix}%
\usepackage{xcolor}%
\usepackage{textcomp}%
\usepackage{manyfoot}%
\usepackage{booktabs}%
\usepackage{algorithm}%
\usepackage{algorithmicx}%
\usepackage{algpseudocode}%
\usepackage{listings}%
\usepackage{subcaption}%
\usepackage{csquotes}
\usepackage{xcolor}

\begin{document}

\title[Article Title]{Reinforcement Learning-Based Laser Cutting Machine Parameter Optimization}


\author[1]{\fnm{Khanh Quan} \sur{Pham}}\email{khanhquan@cbnu.ac.kr}
\author[1]{\fnm{Majid} \sur{Kundroo}}\email{kundroomajid@cbnu.ac.kr}
\author[2]{\fnm{Geunwoo} \sur{Ban}}\email{gwban@npstech.co.kr}
\author[2]{\fnm{Seongho} \sur{Bae}}\email{bsh@npstech.co.kr}
\author*[1]{\fnm{Taehong} \sur{Kim}}\email{taehongkim@cbnu.ac.kr}

\affil[1]{\orgdiv{School of Information and Communication Engineering}, \orgname{Chungbuk National University}, \city{Cheongju}, \postcode{28644}, \country{Korea}}

\affil[2]{\orgdiv{NPS CO., LTD.}, \city{Cheongju}, \postcode{28371}, \country{Korea}}


\abstract{Achieving high accuracy in laser-based cutting of optical films requires careful tuning of parameters such as focal length and laser power beam, adjusted according to the specific properties of each film type. Trial-and-error based traditional methods are used to find the most suitable cutting parameters for various films, but they are slow and inaccurate.
To address this issue, this paper presents the Reinforcement Learning for Laser Cutting (RL$^{2}$C) algorithm, which uses Q-learning with an epsilon-greedy policy to dynamically optimize cutting parameters, significantly reducing taper size and film wastage. Additionally, RL$^{2}$C incorporates a dynamic environment space adaptability mechanism to allow it to adapt to new states encountered during the learning process over multiple batches of experiments.
Experimental results demonstrate that RL$^{2}$C requires fewer steps and less time to find optimal cutting parameters compared to various RL-based optimization methods. Specifically, RL$^{2}$C reduces the number of optimization steps by up to 12.5\% and processing time by up to 81.8\% compared to existing methods. This study demonstrates the potential of RL in industrial laser-cutting processes by improving cut quality, reducing time and film wastage, and minimizing manual interventions.}




\maketitle

\section{Introduction}\label{sec1}
Laser cutting is one of the key enabling technologies in smart manufacturing, which has gained prominence due to its accuracy and efficiency in manufacturing parts for different applications across industries like electronics, automotive aerospace industries, and others.
This technology is capable of cutting many materials, such as metals, plastics, woods, fabrics, polymers, composites, etc., with less wastage and high accuracy; that is why it is essential in those sectors where high accuracy and complex cutting operations are required \citep{Genna2020, SINGH20221021}.
The overall quality of laser cutting depends mainly on key factors such as laser power, pulse frequency, cutting speed, and focus position \citep{GHANY2005438}. Especially, the process of selecting the optimal cutting parameters, particularly laser power and focal length, becomes even more challenging due to the growing variety of materials and cutting requirements. For instance, optical films, which can be used in foldable screens and automotive displays, require fine-tuning of parameters to obtain the optimal cutting quality with the minimum number of defects and unnecessary wastage \citep{chan2016characterization, mi11030248}.

The conventional methods used in the process of identifying the best laser cutting parameters are manual trial-and-error methods and simulation models, which are both time-consuming and inaccurate. These simulation models often use numerical techniques such as finite element analysis (FEA) and computational fluid dynamics (CFD) for estimating thermo-mechanical properties during cutting operations \citep{sundar2013}.
Nevertheless, these methods do not consider the relationships of input and output of cutting parameters, especially in dynamic manufacturing environments \citep{peirovi2017}. To overcome these challenges, some more sophisticated machine learning (ML) algorithms \citep{urgun2024optimization, ren2023modeling} have recently been used to automate the optimization process: artificial neural networks (ANNs), metaheuristic algorithms including genetic algorithms, or particle swarm optimization (PSO). 
Despite the promising nature of the aforementioned approaches, they have not proven effective in exploring huge parameter spaces, working with a wide range of materials, and managing complex industrial environments \citep{urgun2024optimization}. In particular, ANN-based models have high overfitting issues while modeling new or unknown materials, and metaheuristic algorithms face difficulties in converging in real-time environments \citep{ren2023modeling}.

Among the ML approaches, the application of reinforcement learning (RL) \citep{wahab2025}, especially Q-learning, offers a promising solution for automating the work of laser cutting. Unlike supervised learning or metaheuristic algorithms, RL enables an agent to learn from its interaction with the environment and improves its actions based on the rewards gained during the process.
Specifically, Q-Learning can be employed to optimize laser cutting parameters, minimizing the need for manual intervention. The use of RL enables more flexible, continuous adaptation to new materials and conditions as they arise, offering a distinct advantage over static optimization techniques such as Bayesian optimization \citep{WAHAB2020609} or PSO \citep{pramanik2022experimental}.

This paper introduces the RL for laser cutting (RL$^{2}$C) algorithm, a Q-learning-based method designed to optimize laser cutting parameters across three critical stages: focus, power, and taper. RL$^{2}$C enables the agent to update its knowledge over and over as it meets more materials and cutting conditions. 
This adaptive mechanism enhances the algorithm's efficiency and effectiveness, overcoming the shortcomings of the static optimization methods and enabling real-time decision-making in industrial settings. While the core methodology is based on standard tabular Q-Learning, our approach features innovations such as stage-wise optimization and dynamic state-space updates, enabling the algorithm to address the challenges posed by different optical films and their unique properties.

The key contributions of this paper are: 
\begin{itemize} 
\item \textbf{RL$^{2}$C Algorithm:} We propose a novel Q-Learning-based algorithm designed to optimize laser cutting parameters for optical films, integrating a dynamic environment space adaptability mechanism to accommodate real-time learning and the complexities of material characteristics.

\item \textbf{Industrial Application: } This study highlights the practical application of the RL$^{2}$C algorithm in an industrial setting, specifically for laser cutting machines, showcasing its potential to enhance production efficiency and cutting quality. 

\item \textbf{Comparative Analysis:} We present a thorough comparison between RL$^{2}$C and baseline optimization techniques, such as RL-based Bayesian Optimization, RL-based PSO, and Random Search. The analysis demonstrates the effectiveness of our method in time and step efficiency across multiple optimization stages. It highlights the superior performance of RL$^{2}$C in achieving faster convergence and higher precision in parameter selection.
\end{itemize}

The rest of this paper is organized as follows: Section \ref{section:2} discusses the related works, providing an overview of existing optimization techniques in laser cutting, including ML and RL approaches. Section \ref{section:3} presents the overview of the laser cutting process, highlighting material film structures, dataset characteristics, and the optimization stages. Section \ref{section:4} introduces the proposed RL$^{2}$C algorithm, detailing its structure, dynamic adaptability mechanism, and implementation steps. Section \ref{section:5} presents experimental results, including a comparative analysis of RL$^{2}$C with baseline methods in terms of step and time efficiency. Finally, Section \ref{section:6} concludes the study, summarizing key findings and discussing potential future research directions.

\section{Related works}
\label{section:2}
Laser cutting is a widely adopted manufacturing process, but determining the optimal machine parameters remains a significant challenge due to the complex interactions between parameters such as focal length, power, cutting speed, and the material properties of the workpiece \citep{Behbahani_2023}. Traditional approaches, including trial-and-error experiments and simulation-based methods, have notable limitations. These methods are often time-consuming, computationally expensive, and lack the ability to model non-linear relationships or adapt to dynamic changes in the cutting environment \citep{Xu2014}.

ML techniques \citep{liu2023review}, particularly ANNs, have emerged as powerful tools for optimizing laser cutting parameters. ANNs are capable of learning complex, non-linear relationships between input parameters and output quality metrics, such as line width, brightness, and taper size. Early studies by \citep{guo_dixin_laser_2006} and \citep{chen2005influence} used ANNs to quantitatively describe the relationship between cutting quality and parameters in non-vertical laser cutting scenarios. Building on this foundation, \cite{tsai_optimal_2008} integrated ANNs with genetic algorithms, enabling the automatic discovery of optimal parameter settings and improving cutting quality. Beyond ANNs, hybrid ML approaches \citep{park2022machine, wang2025traditional, mi2023situ} have also been explored. \cite{tercan_improving_2017} proposed a clustering and classification framework for laser cutting planning. By first clustering simulation data into performance-based groups and applying classification trees, they identified parameter regions that maximized cutting efficiency. However, these methods often suffer from computational overhead and reduced scalability when applied to high-dimensional parameter spaces or real-time scenarios.

In addition to ML, metaheuristic optimization methods have been applied to parameter optimization for laser cutting. Techniques such as genetic algorithms, PSO, whale optimization algorithm (WOA), and ant lion optimization (ALO) metaheuristic methods have been utilized to explore the parameter space and identify optimal configurations. \cite{urgun2024optimization} demonstrated the effectiveness of these algorithms in optimizing laser cutter parameters. While metaheuristic methods provide robust solutions for static problems, they struggle to adapt dynamically to changing environments or materials, limiting their practicality in real-world applications.

Recent research has shifted toward RL as a more flexible and scalable solution for parameter optimization. RL distinguishes itself from supervised learning (SL) or unsupervised learning (USL) by training through dynamic interactions with the environment rather than relying on static datasets. This makes RL more flexible, as it can adjust its reward function to suit specific tasks and optimize parameters more effectively. \cite{kuprikov2022deep} employed deep RL to optimize laser cutting parameters, showcasing its ability to achieve higher precision compared to static optimization methods. Similarly, \cite{Behbahani_2023} applied ML techniques, including RL, to optimize the machining process of alumina ceramics. Besides, \cite{chang2024optimization} explores the optimization of laser annealing parameters using Bayesian RL (BRL). This study leverages fixed and variable prior knowledge derived from experimental data and technology computer-aided design (TCAD) simulations to guide the RL agent in selecting optimal parameters, such as laser power, repetition rate, and processing temperature. \cite{zhang2023q} proposed a Q-learning-based multi-objective PSO (QL-MoPSO) framework for distributed flow-shop scheduling problems. Their hybrid approach balanced exploration and exploitation capabilities, achieving faster convergence and improved diversity in solutions.

However, existing RL approaches often assume deterministic state transitions and static learning mechanisms. This means that if a state is not included in the predefined state space during training, the RL agent may struggle to generalize and adapt effectively when encountering unseen conditions. Building on this foundation, our proposed RL$^{2}$C algorithm addresses these limitations by introducing a dynamic environment space adaptability mechanism that continuously adapts to new material conditions. Unlike static RL models, RL$^{2}$C updates the Q-table in real time for each environment, allowing the agent to effectively explore and exploit the parameter space across diverse materials and cutting conditions. 

\section{Overview of Laser Cutting Process}
\label{section:3}
This section presents a comprehensive overview of the material film structure, the configuration of the input dataset used for training, the measurement techniques used to evaluate the cutting quality, and the formulation of optimization problems.

\subsection{Material Film Structure and Input Dataset Configuration}\label{subsec3.1}
\begin{figure*}[h]
\centering
\includegraphics[width=\textwidth]{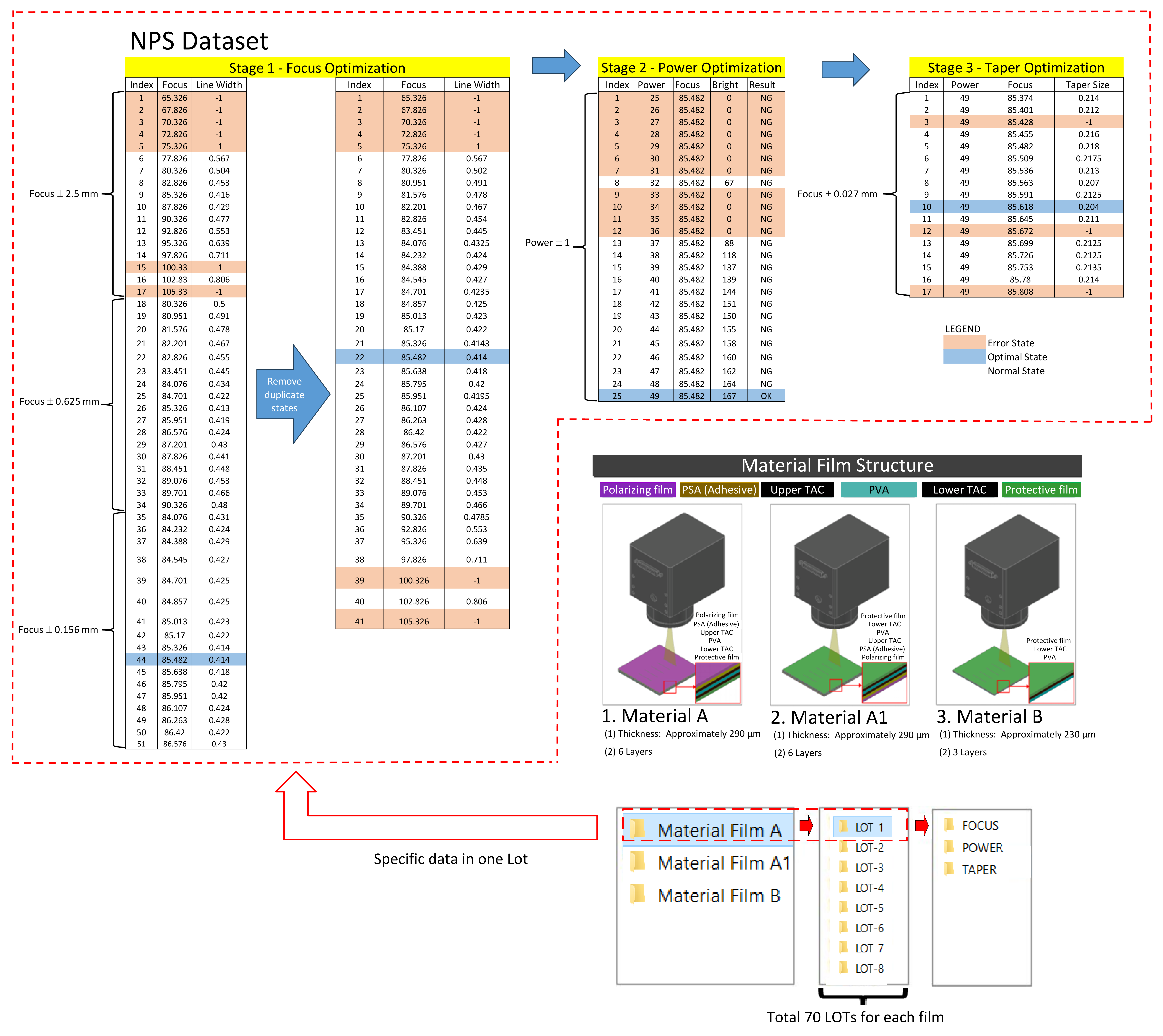}
\caption{Material film structure and input dataset configuration for RL$^{2}$C training.}\label{fig:nps_architecture}
\end{figure*}

Fig. \ref{fig:nps_architecture} illustrates the structure of the material films and the corresponding input dataset configuration used for training the RL$^{2}$C algorithm.

The bottom part of Fig. \ref{fig:nps_architecture} presents the data characteristics grouped according to the material. There are three types of material films: Material A, Material A1, and Material B, with each material film having 70 lots for data collection. Each lot represents a unique set of experimental conditions applied to that specific material film. Within each lot, the data are further categorized into three subfolders: \textbf{focus}, \textbf{power}, and \textbf{taper} which correspond to the three optimization stages. These subfolders store the raw measurement and parameter-tuning data associated with each stage of the laser-cutting process.

To clarify how diverse material films influence the learning behavior of RL$^{2}$C, the middle part of Fig. \ref{fig:nps_architecture} outlines the physical layer structure of each material film. Material A and A1 consist of six layers, including a polarizing film, Pressure-Sensitive Adhesive (PSA), Upper Tri-Acetyl Cellulose (TAC), Polyvinyl Alcohol (PVA), Lower TAC, and a protective film, with a total thickness of approximately 290 $\mu$m. Material B, on the other hand, has a simpler structure, consisting of three layers, including a protective film, lower TAC, and PVA, with a total thickness of approximately 230 $\mu$m.

The upper part of Fig. \ref{fig:nps_architecture} represents the sequential flow of the dataset across the three optimization stages—\textbf{Focus Optimization}, \textbf{Power Optimization}, and \textbf{Taper Optimization}. It provides a detailed example of data collected from a single lot, showcasing the step-by-step progression through each stage of the laser-cutting process. Each row in this upper section corresponds to a trial conducted under a specific parameter setting, while the columns capture both the control variables (e.g., focus value or power value) and the resulting measurements (e.g., line width, brightness, or taper size) for that stage.
It is important to note that the upper part of Fig. \ref{fig:nps_architecture} illustrates the configuration and results from one representative lot only; the same structure applies to all 70 lots per material film, but the specific parameter values and measurements differ depending on the material characteristics and experimental variations. These configurations are directly used in the learning process of RL$^{2}$C. Each row in the dataset serves as a state-action-reward tuple, which forms the basis for Q-learning updates across episodes. 
Based on this data structure, the specific procedures and data handling approaches for each optimization stage are detailed below, beginning with Stage 1.

In Stage 1, Focus Optimization aims to identify the focus value that minimizes the line width. By adjusting the focus value with specific variation sizes ($\pm$2.5, $\pm$0.625, $\pm$0.156 mm), the process begins with selecting a range of focus values and testing them iteratively. Then we measure and record the line width for each focus value setting. During this process, each focus value is categorized into states: error state for invalid or failed attempts (e.g., line width = -1), normal state for acceptable but suboptimal results, and optimal state for the minimum line width value. When duplicate focus values are present, their corresponding line width values are computed as the mean to ensure the data are accurate and reliable. For instance, at the end of Stage 1, the optimal focus setting identified is 85.482, resulting in a minimum line width of 0.414.

In Stage 2, by using the optimal focus value obtained from Stage 1 as a fixed input parameter, Power Optimization aims to determine the power value that ensures the laser cut meets a specific brightness threshold. The process involves adjusting the power incrementally (e.g., $\pm$1\%) and measuring the resulting brightness for each setting. Each power value is tested while keeping the focus constant at the optimal setting identified in Stage 1. We measure and record the brightness, and the power value that achieves the desired brightness is identified as the optimal state (e.g., result = OK). Error states whose brightness equals zero and the result is marked as \enquote{NG} (not good) (e.g., brightness = 0 and result = NG), while normal states are the remaining ones (e.g., brightness $\neq$ 0 and result = NG). The optimal power value of 49 identified in Stage 2 will be used in the final stage of the optimization process. 

\begin{figure*}[h]
  \centering
  \includegraphics[width=\linewidth]{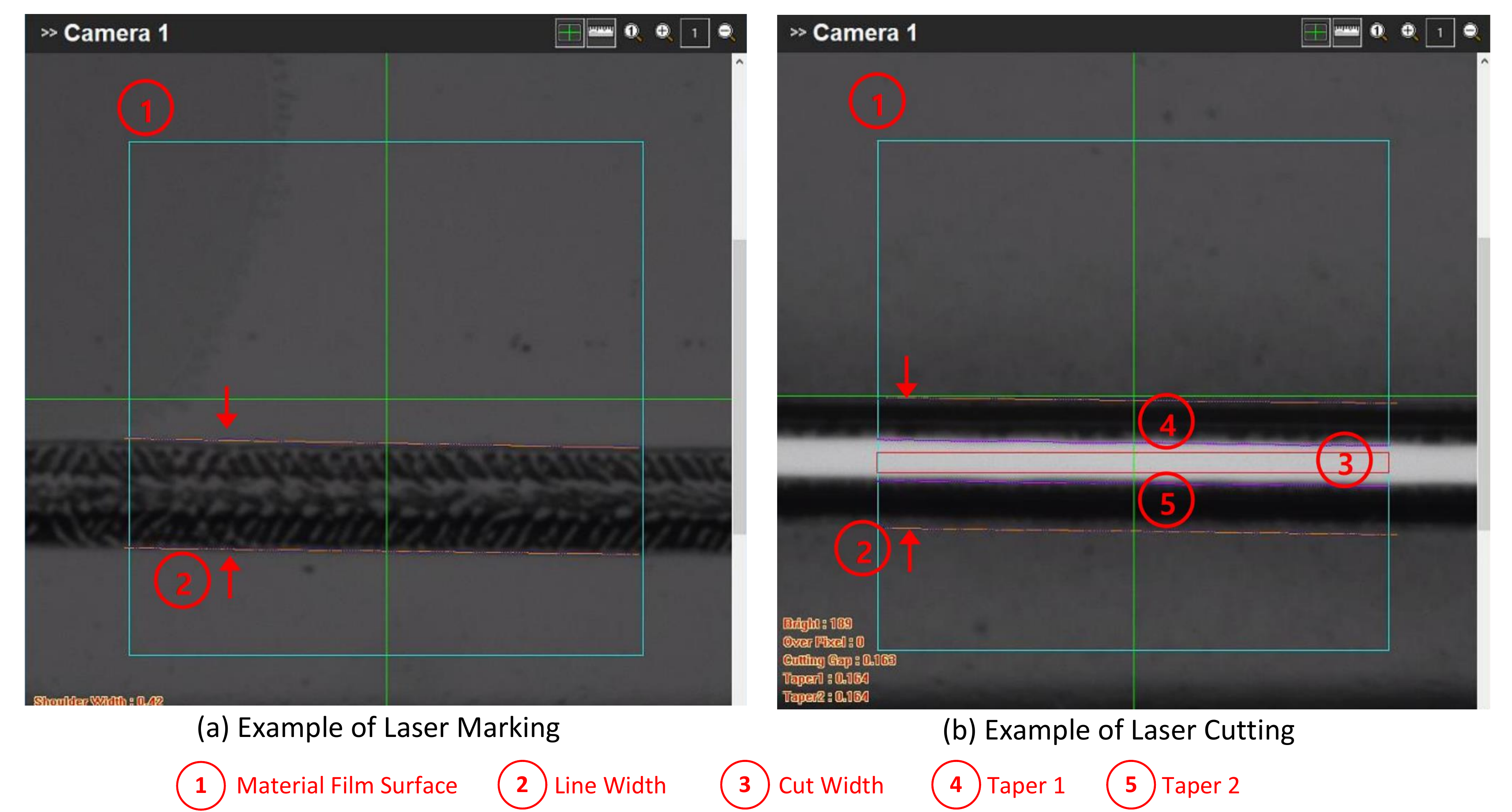}
  \caption{Examples of laser marking and cutting.}
  \label{fig:laser}
\end{figure*}

In Stage 3, Taper Optimization focuses on minimizing the taper size. Using the optimal focus 85.482 mm and power value of 49 determined from Stages 1 and 2, the focus is fine-tuned within a narrow range (e.g., optimal focus from Stage 1 $\pm$0.027 mm) to minimize the taper size. Similar to the previous stages, the focus value is incrementally adjusted, and the resulting taper size is measured for each setting. Moreover, each focus value is also categorized into states: error state for taper size = -1, optimal state for the smallest taper size, and normal state for the remaining one. At the end of Stage 3, the optimal focus setting (85.618 mm) that minimizes the taper size is identified and used for production, completing the optimization process.

This three-stage structure forms a consistent learning workflow across all lots and materials. Importantly, the entire pipeline—from data collection to learning—is designed to operate in a fully autonomous manner, with minimal human intervention. Human involvement is limited to loading the film material into the laser cutting machine and specifying the number of lots to be processed. Once initiated, the machine automatically executes all cutting procedures sequentially across the three optimization stages—starting with focus, followed by power, and then taper optimization. For each parameter configuration, the system measures the corresponding cutting quality (e.g., line width, brightness, taper size) and stores the results in structured CSV files. These files are formatted in the same structure as shown in the upper part of Fig. \ref{fig:nps_architecture}, including both the parameter values and their measured outcomes.

Once the generated CSV files from all lots are ready and the environment is initialized, the remaining stages—from state exploration and action selection to Q-table updates and policy refinement—are fully automated. The RL$^{2}$C agent autonomously interacts with the environment, optimizes parameters across episodes, and dynamically adapts to new states using its dynamic Q-learning mechanism. This division of responsibilities ensures that human oversight remains minimal and is only required during the initial experimental setup. After that, the learning and decision-making processes operate in an autonomous and self-improving manner. This makes the RL$^{2}$C framework both scalable and practical for real-world manufacturing environments.

\subsection{Laser Marking and Cutting Quality Measurements}\label{subsec3.2}

Fig. \ref{fig:laser} presents examples of laser marking and cutting, highlighting key measurement points for assessing the quality of the laser cutting process. In Fig. \ref{fig:laser}\textcolor{blue}{a}, the surface of the material film \textbf{\textcircled{1}} is displayed, with the line width \textbf{\textcircled{2}} being measured to evaluate focus optimization. This line width represents the thickness of the black line on the material surface, and it is a critical parameter for ensuring precise marking.

In Fig. \ref{fig:laser}\textcolor{blue}{b}, several measurements are shown to optimize power and taper. The cut width \textbf{\textcircled{3}} measures the thickness of the white region in the middle, representing the fully cut area. Taper 1 \textbf{\textcircled{4}} and Taper 2 \textbf{\textcircled{5}} refer to the black regions on either side of the cut. The narrower the black region, the better cutting quality. Since taper size can be asymmetric, the sum of Taper 1 and Taper 2 is averaged to determine the overall cut quality.

Several output parameters define the quality of the cut, which are line width, brightness, and taper size. Line width is related to the width of the cutting line made by the laser beam, and it determines the accuracy of the cutting. Brightness is the quality of the cut surface and the intensity of the dark spots left on the material after the cut. Taper size is the difference in the width of the cut between the top and bottom surfaces of the material, affecting the consistency of the cut. The above-mentioned output parameters are controlled to achieve high-quality cutting, although several input parameters like focus and power can affect them. Focus or focal length refers to the distance between the focusing lens and the material being cut, while power refers to the intensity of the laser beam.

\subsection{Problem formulation}
The optimization of the laser cutting process involves three key stages, each targeting a specific parameter to improve cutting quality. The problem is formulated as a sequential optimization process, where each stage refines a parameter to achieve the best results under given constraints.

\subsubsection{Stage 1: Focus Optimization}
Focus Optimization aims to minimize the line width, $L_w$, which is a function of the focus value, $F$. The objective function for this stage is:
\begin{equation}
    \min_{F} L_w(F)
\end{equation}
where $F$ represents the focus value. The constraints for this stage are:
\begin{itemize}
    \item $F \in \{F_{\text{start}}, F_{\text{start}} \pm \Delta F, \ldots, F_{\text{end}}\}$, where $\Delta F = \pm 2.5$, $\pm 0.625$, $\pm 0.156 \, \text{mm}$.
    \item $L_w(F) > 0$ and $L_w(F) \neq -1$ (valid line width values).
    \item Duplicate focus values are removed, and their corresponding line width values are averaged.
\end{itemize}

\subsubsection{Stage 2: Power Optimization}
Power Optimization determines the laser power, $P$, that meets a specific brightness threshold $B_{\text{thresh}}$. The optimization objective is expressed as:
\begin{equation}
    \min_{P} \left( B(P)-B_{\text{thresh}} \right) \textit{subject to } B(P)\geq B_{\text{thresh}}
\end{equation}

The constraints for this stage include:
\begin{itemize}
    \item $P \in \{P_{\text{start}}, P_{\text{start}} \pm \Delta P, \ldots, P_{\text{end}}\}$, where $\Delta P = \pm 1 \, \%$.
    \item Only power values where $B(P) \geq B_{\text{thresh}}$ are valid.
    \item Result $(P) = \text{OK}$ for optimal states.
\end{itemize}

\subsubsection{Stage 3: Taper Optimization}
Taper Optimization minimizes the taper size, $T_s$, using the optimal focus and power values from the previous stages. The optimization objective is:
\begin{equation}
    \min_{F} T_s(F_{\text{opt}}, P_{\text{opt}})
\end{equation}
with the constraints:
\begin{itemize}
    \item $F \in \{F_{\text{opt}} \pm \Delta F, \ldots, F_{\text{opt}} + n \cdot \Delta F\}$, where $\Delta F = \pm 0.027 \, \text{mm}$.
    \item $T_s(F_{\text{opt}}, P_{\text{opt}}) > 0$ and $T_s(F_{\text{opt}}, P_{\text{opt}}) \neq -1$ (valid taper size values).
\end{itemize}

\begin{figure*}[!h]
  \centering
  \includegraphics[width=0.8\linewidth]{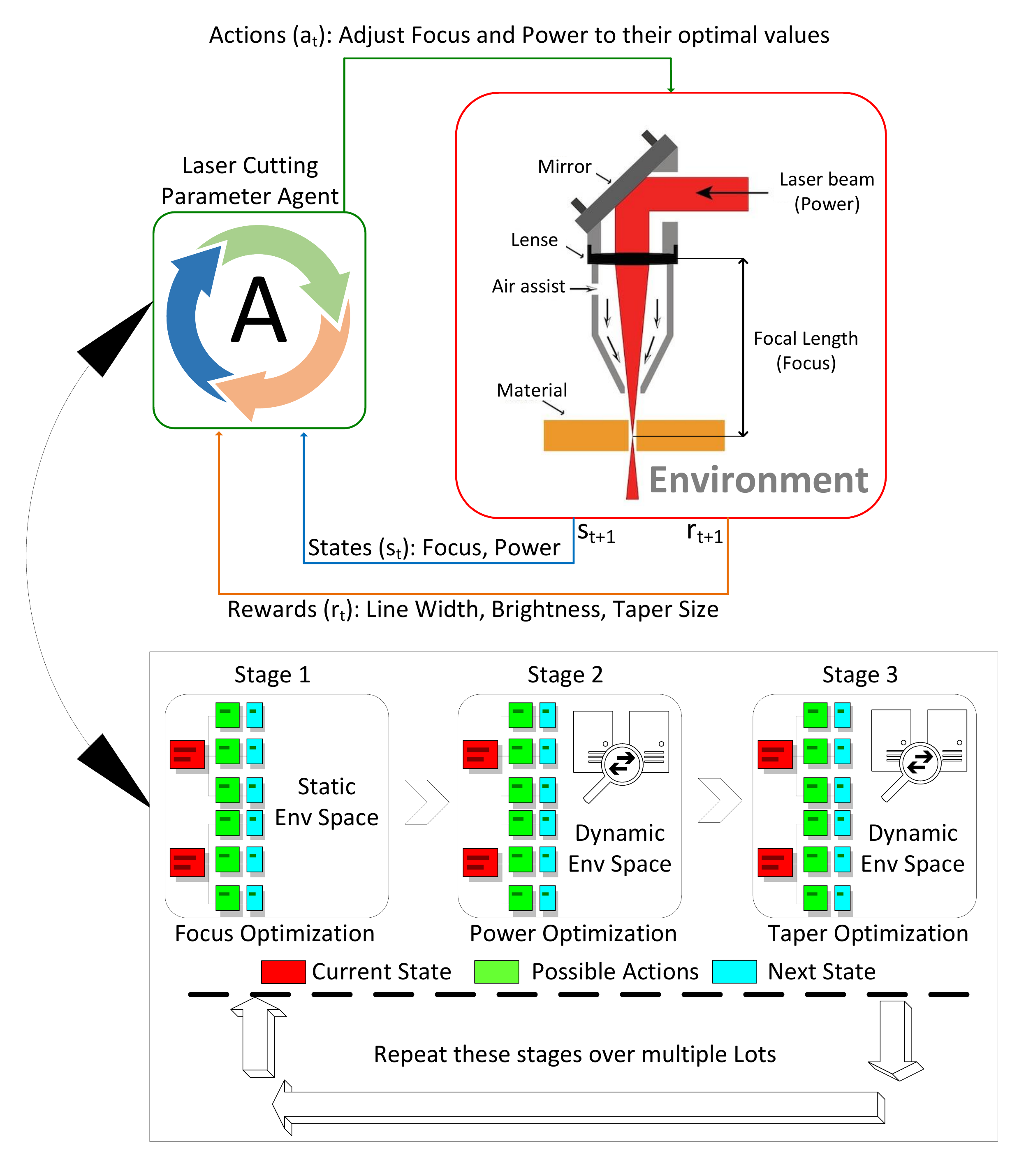}
  \caption{Reinforcement learning framework for laser cutting (RL$^{2}$C) parameter optimization.}
  \label{fig:architecture}
\end{figure*}

\subsubsection{Overall Objective}
The overall optimization can be formulated as a multi-stage process, where the combined objective is:
\begin{equation}
\hspace*{-0.25cm}
\min_{F, P} \left\{ L_w(F), (B(P) - B_{\text{thresh}}), T_s(F_{\text{opt}}, P_{\text{opt}}) \right\}
\end{equation}
subject to the constraints defined for each stage.

The three optimization stages above are designed to operate sequentially, with each stage building upon the outcome of the previous one. Specifically, Stage 2 takes the optimal focus value $F_{\text{opt}}$ obtained from Stage 1 as a fixed input to determine the corresponding optimal power value $P_{\text{opt}}$. Stage 3 then uses both $F_{\text{opt}}$ and $P_{\text{opt}}$ as inputs to optimize the taper size. Importantly, Stage 3 does not explore the entire focus search space evaluated in Stage 1; instead, it performs a search within a narrow range centered around $F_{\text{opt}}$ (e.g.,$F_{\text{opt}} \pm \Delta F$). This design makes Stage 3 inherently dependent on the results of both Stage 1 and Stage 2, while also effectively refining the parameter found in Stage 1 to achieve the best possible final quality.
\section{Proposed Method}
\label{section:4}
\subsection{Laser Cutting Parameter Optimization Agent}

Laser-cutting machines are precision-cutting instruments that utilize high-power laser beams on the material that is to be cut. The quality of the cut is primarily determined by output parameters such as line width, brightness, and taper size. These parameters must be optimized to ensure optimal cutting quality. Among the key factors influencing these outputs are the input parameters, including focus and power, which directly control the cutting process. 

Fig.~\ref{fig:architecture} presents the architecture of the RL$^2$C framework, which is trained using the structured dataset illustrated in Fig. \ref{fig:nps_architecture}. In this framework, the laser-cutting parameter agent interacts with an environment that simulates the laser-cutting process. The environment receives focus and power values as actions ($a_t$) and returns the resulting cutting quality metrics: line width, brightness, and taper size as rewards ($r_t$), along with the next state ($s_{t+1}$). This process is structured into three stages: Stage 1 (Focus Optimization), Stage 2 (Power Optimization), and Stage 3 (Taper Optimization).

% Fig. \ref{fig:architecture} describes the architecture of our RL framework used for optimizing laser cutting parameters, specifically focus and power settings. Essentially, a laser cutting parameter agent communicates with its surroundings; this includes the laser cutting system which consists of a laser beam (power), the material on which the laser is directed, as well as the focus (focal length) adjustments. The agent receives states ($s_{t}$) related to focus and power, performs actions ($a_{t}$) to adjust these parameters, and receives rewards ($r_{t}$) based on the quality metrics like line width, brightness, and taper size. 

In Stage 1, the agent operates within a static environment space, where the state space remains consistent across all lots. This stage focuses on optimizing the focus parameter under stable conditions, allowing the agent to refine the focus settings effectively. 

From Stage 2 to Stage 3, the environment turns into a dynamic one, meaning that the state space varies over lots, presenting new challenges for the agent. In Stage 2, the dynamic environment arises because the optimal power setting required to achieve the desired brightness depends on material-specific characteristics, such as variations in composition and thickness across lots. The agent must adjust the power parameter to account for these differences dynamically, ensuring that the brightness threshold is satisfied across lots.

\begin{algorithm*}[!ht]
    \caption{RL$^{2}$C Algorithm}
    \label{alg:1}
    \begin{algorithmic}[1]
        \Statex {\textbf{Input:} $\alpha$ = 0.9, $\gamma$ = 0.9, $\varepsilon_{max}$ = 1.0, $\varepsilon_{min}$ = 0.1, $decay\_rate$ = 0.05.}
        \Statex {\textbf{Output:} Optimal state $s^{*}$, optimal policy $\pi^{*}$, optimal Q-table $Q^{*}$.}
        \Procedure{RL AGENT}{} 
            \State{Initialize Q-Table $Q$ for all state-action pairs.}
            \For{each $material$} 
                \For{each $lot$}
                    \State{$\varepsilon \leftarrow \varepsilon_{min} + (\varepsilon_{max}-\varepsilon_{min})e^{-decay\_rate*lot}$}
                    \For{each $episode$}
                    \State{$s'$, $r$ $\leftarrow$ Epsilon\_Greedy\_Policy($\varepsilon$, $s$)}
                    
                    \State{Update the Q-values for current state-action pairs using the Bellman Equation in Eq. \ref{bellman_equation}.}
        
                    % \State{$s \leftarrow s'$}
                    %     \If{$s'$ == $s^{*}$ \textbf{or} $s'$ == Error state}
                    %         \State{\textbf{break}}
                    %     \EndIf
                    \State{$s \leftarrow s'$; \textbf{break} if $s'$ == $s^{*}$ \textbf{or} $s'$ == Error state}
                    \EndFor
                \State{Update optimal policy $\pi^{*}(s) \leftarrow \underset{a}{\arg\max}(Q[s, a])$.}
                \State{\colorbox{lime}{$Q^{*}$ $\leftarrow$ UPDATE\_STATIC\_Q\_TABLE($Q$) \Comment{Stage 1.}}}
                \State{\colorbox{cyan!20}{$Q^{*}$ $\leftarrow$ UPDATE\_DYNAMIC\_Q\_TABLE($Q$) \Comment{Stage 2 and 3.}}}

                \EndFor
            \EndFor
            \State{\textbf{return} $\pi^{*}$, $Q^{*}$}
        \EndProcedure
        \Procedure{Epsilon\_Greedy\_Policy}{$\varepsilon$, $s$} 
            \If{$random\_uniform(0,1) < \varepsilon$}
                \State{$a \leftarrow random\_action()$}
            \Else
                \State{$a \leftarrow \underset{a}{\arg\max}(Q[s, a])$}
            \EndIf
            \State{$s'$ $\leftarrow$ $s$ + $a$}
            \State{$r$ $\leftarrow$ Reward\_Function($s$, $a$, $s'$, $s^{*}$) \Comment{Eq. \ref{stage1_reward_function} for Stage 1, Eq. \ref{stage2_reward_function} for Stage 2, and Eq. \ref{stage3_reward_function} for Stage 3.}}
            \State{\textbf{return} $s'$, $r$}
        \EndProcedure

        \Procedure{Update\_Static\_Q\_table}{$Q$}
            \State{A new static Q-table is updated after each lot using Eq. \ref{bellman_equation}.}
        \EndProcedure
        
        \Procedure{Update\_Dynamic\_Q\_table}{$Q$}
        \If{new states exist}
            \For{each new state}
                \State{Compute possible next states for new states.}
                \State{Add more rows in $Q$ for new states.}
            \EndFor
            \For{each existing state that can reach new states}
                \State{Update $Q$ values where transitions are possible.}
            \EndFor
        \EndIf
        \State{\textbf{return} $Q^{*}$}
        \EndProcedure
    \end{algorithmic}
\end{algorithm*}

Similarly, Stage 3 also operates in a dynamic environment, as taper size is influenced by both material properties and the cumulative effects of focus and power parameters selected in the earlier stages. This stage requires the agent to fine-tune the focus setting chosen in Stage 1 while using power adjustments made in Stage 2, as well as material-specific variations in taper behavior. The evolving state space in these stages reflects real-world variability, pushing the agent to adapt its actions dynamically for each lot.

Throughout these stages, the agent learns optimal parameter settings across multiple lots, adjusting its actions based on both static and dynamic environments. This dynamic adaptability is key to achieving high-quality laser cutting in real-world applications, where material properties and conditions can vary significantly across production lots.

%%%%%%%%%%%%%%%%%%%%%%%%%%%%%%%%%%%%%%%%%%%%%%%%%%%%%%%%%%%%%%%%%%%%%%%%%%%%%%%%%
\subsection{RL$^{2}$C Algorithm}\label{sec4}

The RL$^{2}$C in Algorithm \ref{alg:1} is designed to optimize laser cutting parameters such as focus and power. It uses Q-learning to train an agent to identify the best setting for high-quality cutting for various material films and lots. In this subsection, the details of the algorithm implementation are discussed, including input and output parameters, the iterative learning process, and its capability to adjust to the static and dynamic changes in the environment during various phases of the optimization.

The RL agent takes as input a set of hyperparameters \citep{majid2023}: the learning rate $\alpha$, the discount factor $\gamma$, the exploration-exploitation rate $\epsilon$, and the rate of decay of the exploration parameter. The output of the algorithm is the optimal state $s^{*}$, the optimal policy $\pi^{*}$, and the optimal Q-table $Q^{*}$, which contains the best laser cutting settings with the corresponding chain of actions required to achieve them.

The RL$^{2}$C algorithm begins with the initialization of the environment, which determines the states and actions of the laser-cutting process. At this time, the Q-table is populated with zero values or small random positive values for each state-action pair. The agent also initializes its policy $\pi^{*}$, which will guide its actions based on the Q-values. The agent is now ready to explore and interact with the environment.

Each Q-table is structured as a key-value map, where each key is a state-action pair, and the value is the calculated Q-value. The state typically is current parameter values (e.g., focus or power), and the action represents discrete steps to increment or decrement these parameters. This design enables flexible learning across multiple lots, with the agent progressively refining its knowledge in each stage without interference from others.
Once the environment and Q-table are set up, the agent starts its learning process when testing with many different materials. For each lot, the agent must decide whether to explore new actions or exploit the best-known actions based on its current knowledge. This decision is made using the $\varepsilon$-greedy policy, responsible for managing the trade-off between exploration and exploitation. At each step, the agent performs an action and obtains a response from the environment in the form of a reward function, which is calculated using line width, brightness, and taper size in Eq. \ref{stage1_reward_function}, Eq. \ref{stage2_reward_function}, and Eq. \ref{stage3_reward_function}, respectively.

After each action, the Q-values are updated using the Bellman equation: 

\begin{equation}
\label{bellman_equation}
Q(s, a) \leftarrow (1 - \alpha) Q(s, a) + \alpha \left[ r + \gamma \max_{a'} Q(s', a') \right]
\end{equation}

This equation adjusts the Q-value for the state-action pair based on both the immediate reward and the expected future rewards from the next state. After each episode, the agent updates its policy $\pi^{*}$, selecting actions that maximize its expected rewards for each state. The agent repeats this process until it reaches the optimal state $s^{*}$, encounters an error state, or gets the maximum number of steps to be performed in one episode. This process helps the agent better understand which of the possible optimal actions will result in the highest reward.

The RL$^{2}$C algorithm is developed to work for both static and dynamic environments, which represent the different conditions of the laser cutting process. One of the significant characteristics of this algorithm is the Update\_Dynamic\_Q\_table() mechanism with which the agent learns to modify the Q-table in response to new states during training across multiple lots. As new states arise due to changes in material properties or laser cutting conditions, the agent incorporates them into the Q-table by adding new rows and updating transitions from existing states. This ensures that the agent can continue to explore and exploit these newly discovered states, maintaining optimal performance in an evolving environment. By doing so, the Q-table is continuously updated to account for these environmental changes, making the RL$^{2}$C algorithm robust and adaptable to the dynamic nature of real-world manufacturing scenarios.

To support stage-wise learning and maintain separation between optimization objectives, RL$^{2}$C maintains a separate Q-table for each stage: one for focus optimization (Stage 1), one for power optimization (Stage 2), and one for taper optimization (Stage 3). The Q-table in Stage 1 operates in a static environment, where the state space is predefined and fixed. In contrast, the Q-tables for Stage 2 and Stage 3 are dynamic; their state spaces evolve based on new input-output observations from each lot, allowing the agent to adapt to changes in material characteristics or cutting conditions. 

Once the learning process is done at all of the stages, it returns to the optimal policy 
$\pi^{*}$ and the final Q-table $Q^{*}$. These represent the agent’s learned strategy for achieving the best laser-cutting results, as well as the expected rewards for each state-action pair. At the end of the process, the agent enhances the knowledge about the environment and identifies the optimal actions for every step taken in all lots.

\subsection{Detailed Design of Each Stage Using RL$^{2}$C}

This subsection describes how to apply the RL$^{2}$C algorithm in each state space, action space, and reward function.

\subsubsection{Stage 1 - Focus Optimization}

\begin{itemize}
    \item \textbf{State Space}: The state space is defined by the current focal length of the laser. Each state represents the system’s configuration at a specific moment during the optimization process, with focus adjustments made across three phases. As shown in Fig. \ref{fig:nps_architecture}, in \textbf{Phase 1}, the focal length starts at an initial value of 65.326 mm and is adjusted by 2.5 mm increments over 17 iterations. After completing 17 steps, the focus value that satisfies the minimum line width among the results is selected.
    In \textbf{Phase 2}, based on the minimum line width value selected in \textbf{Phase 1} scale, optimization is performed again by setting that focus value as the midpoint and incrementing or decrementing by $\pm$0.625 mm in 17 steps. Similarly, the focus value that satisfies the minimum line width is selected. In \textbf{Phase 3}, continuing from the last result in \textbf{Phase 2} scale, the final optimization is performed in \textbf{Phase 3} scale with 17-step adjustments with $\pm$0.156 mm. This state space allows the agent to explore how precise focus adjustments impact the system's performance, as it narrows down the optimal focus setting within these specific focus ranges.

    \item \textbf{Action Space}: The available actions include both positive and negative adjustments in millimeters: [$\pm$2.5, $\pm$0.625, $\pm$0.156,$\pm$ 0.157]. These actions allow the agent to either increase or decrease the focal length incrementally, depending on the feedback it receives from the environment.  

    \begin{figure}[!h]
    \centering
    \includegraphics[width=\linewidth]{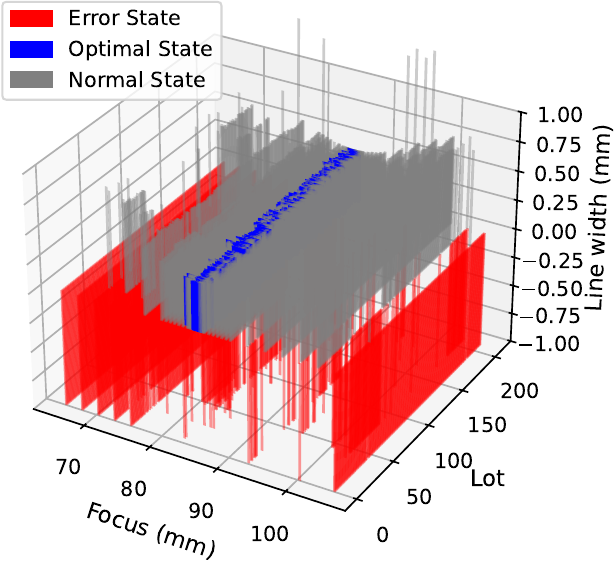}
    \caption{3D parameter space visualization for Stage 1 - Focus Optimization.}
    \label{stage1_param_visual}
    \end{figure}
    
    \item In our approach, each state that the agent encounters can be classified into one of three categories: \textbf{Optimal State}, \textbf{Error State}, or \textbf{Normal State}, based on the current focal length and the corresponding line width result. 
    The \textbf{Optimal State} is the state where the focal length produces the narrowest line width, indicating the best possible performance for the laser marking process. These states are represented by the blue points in the 3D parameter space visualization in Fig. \ref{stage1_param_visual}, which are concentrated near the optimal focal length range (e.g., around 85.5 mm). 
    The \textbf{Error State} occurs when the linewidth cannot be measured properly, typically caused by an incorrect focal length that prevents proper marking (e.g., line width = -1.0). In the 3D parameter space, these states are depicted as red points, scattered outside the optimal range, highlighting regions where the cutting process fails.
    Lastly, \textbf{Normal State} refers to any state where the line width is measurable but not optimal. These states are represented by gray points in the visualization, forming a transition zone between error and optimal states. In these states, the agent receives negative feedback, as the focal length does not produce the narrowest possible line, but the process is still functional.

    \item \textbf{Reward Function}: The reward function guides the agent in selecting the best focal length while penalizing inefficient actions. The reward is defined as:
    \begin{equation}
    \label{stage1_reward_function}
        R1=
        \begin{cases} 
        -200,\textit{if Error}, \\
        \frac{100}{\textit{line width}}-(step\times0.1),\textit{if Optimal}, \\
        -\frac{1}{\textit{line width}}-(step\times0.1),\textit{if Normal}.
        \end{cases}
    \end{equation}
    In Eq. \ref{stage1_reward_function}, lower line width values yield higher rewards, incentivizing more precise cuts. For \textit{error states} (e.g., line width = -1), a fixed penalty of -200 is assigned to strongly discourage invalid settings. For \textit{optimal states}, defined as the minimum line width observed in the dataset (Fig. \ref{fig:nps_architecture}), the reward scales with $\frac{100}{\textit{line width}}$, typically producing values in the range of approximately 125 to 250 for line width values between 0.4 mm and 0.8 mm. The \textit{normal states} yield a smaller reward, ranging from approximately 1.25 to 2.5, calculated as $\frac{1}{\textit{line width}}$. In both \textit{optimal} and \textit{normal states}, the reward is slightly penalized by the number of steps taken in the episode via the $(step \times 0.1)$ term, encouraging faster convergence.
\end{itemize}

\subsubsection{Stage 2 - Power Optimization}
\begin{itemize}
    \item \textbf{State Space}: This is defined by the current power setting and the analysis of the cutting result for the current lot. The initial power setting starts at 25\%, and the power is incremented by 1\% in each step. For each power setting, the system analyzes the cutting image to evaluate the cutting quality. A successful cut is determined based on two conditions: \textbf{(1)} no dark spots are present within the cutting area, and \textbf{(2)} the average brightness value within the cutting area is 167 or higher. The state space thus consists of the power setting at each step, the absence or presence of dark spots, and the average brightness of the cut.

    \item \textbf{Action Space}: [$\pm$10, $\pm$9, $\pm$8, $\pm$7, $\pm$6, $\pm$5, $\pm$4, $\pm$3, $\pm$2, $\pm$1] \%.

    \item In Stage 2, the \textbf{Error State} occurs when the cutting brightness is 0, resulting in a significant penalty. These states are represented by red points in the 3D parameter space visualization in Fig. \ref{stage2_param_visual}, showing regions where the power settings fail to generate sufficient brightness for a valid cut.
    The \textbf{Optimal State} is reached when the cutting brightness matches the optimal value, indicating a successful cut without dark spots, and the average brightness value is higher or equal to a specific threshold. In the parameter space, these states are depicted as blue points, concentrated in a specific range of power values.
    And the \textbf{Normal state} covers any non-optimal, measurable brightness values. These are represented by gray points in the visualization, forming a transition zone between error and optimal states.

  \begin{figure}[!h]
    \centering
    \includegraphics[width=\linewidth]{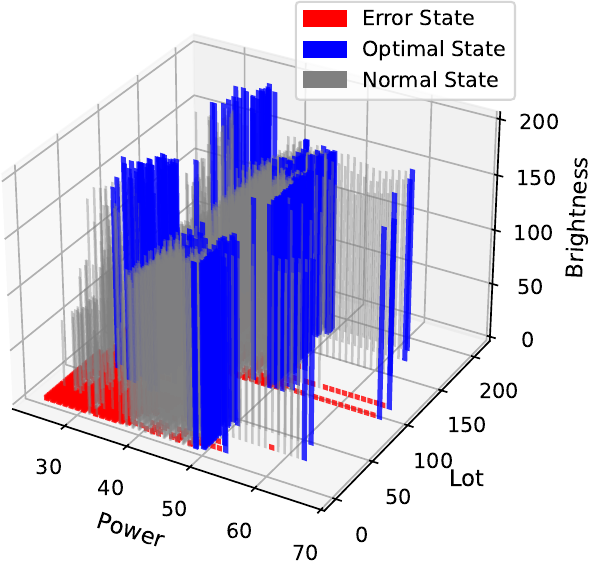}
    \caption{3D parameter space visualization for Stage 2 - Power Optimization.}
    \label{stage2_param_visual}
    \end{figure}

    \item \textbf{Reward Function}: 
    The reward function for Stage 2 is defined as:
    \begin{equation}
        \label{stage2_reward_function}
        R2=
        \begin{cases} 
        -5000, \textit{if Error}, \\
        (\textit{brightness} \times 50) - (\textit{step} \times 2), \textit{if Optimal}, \\
        -\textit{brightness} - (\textit{step} \times 2), \textit{if Normal}.
        \end{cases}
    \end{equation}
    In Eq. \ref{stage2_reward_function}, the reward is primarily influenced by the brightness of the laser mark. According to the dataset in Fig. \ref{fig:nps_architecture}, brightness values range from 0 (\textit{error states}) to a maximum of 167 in the optimal configuration. For \textit{optimal states}, the reward can reach up to \(167 \times 50 = 8350\) before applying the step penalty, making it highly rewarding. For \textit{normal states}, brightness values are non-zero but insufficient to meet the threshold, resulting in negative rewards ranging roughly from -1 to -164. A strong fixed penalty of -5000 is used for error states where brightness is zero. The step penalty term \((\textit{step} \times 2)\) discourages prolonged exploration and encourages faster convergence.
\end{itemize}

\subsubsection{Stage 3 - Taper Optimization}

\begin{itemize}
    \item \textbf{State Space}: This is defined by the current focus and power settings, with better focus adjustments applied to optimize the taper value. Starting with the best focus and power values from the previous stages, the focus is adjusted incrementally by 0.027 mm over 17 iterations. The system evaluates the taper quality by calculating the average of \textbf{(Taper1 + Taper2) / 2}. A successful cut is determined when the taper result meets the same cutting conditions as in Stage 1: the system returns an \enquote{OK} if the taper values are acceptable, indicating a successful cut. The state space consists of the focus adjustments, the calculated taper value, and the result status (\enquote{OK} or \enquote{NG}).

    \item \textbf{Action Space}: [$\pm$0.109, $\pm$0.108, $\pm$0.056, $\pm$0.055, $\pm$0.054, $\pm$0.028, $\pm$0.027] mm.
    
    \item In Stage 3, the \textbf{Error State} occurs when the taper result is invalid (i.e., taper value of -1.0), leading to a penalty. These states are represented by red points in the 3D parameter space visualization in Fig. \ref{stage3_param_visual}, highlighting regions where the focus adjustments fail to produce valid taper measurements.
    The \textbf{Optimal State} is achieved when the taper result produces the minimum \((Taper1 + Taper2) / 2\) value. These states are shown as blue points in the 3D visualization, clustered around specific focus values.
    The \textbf{Normal State} refers to any valid taper value that is not optimal, where the agent continues to adjust focus to reach the optimal taper. These states are depicted as gray points in the visualization, forming a transitional zone between error and optimal states.

  \begin{figure}[!h]
    \centering
    \includegraphics[width=\linewidth]{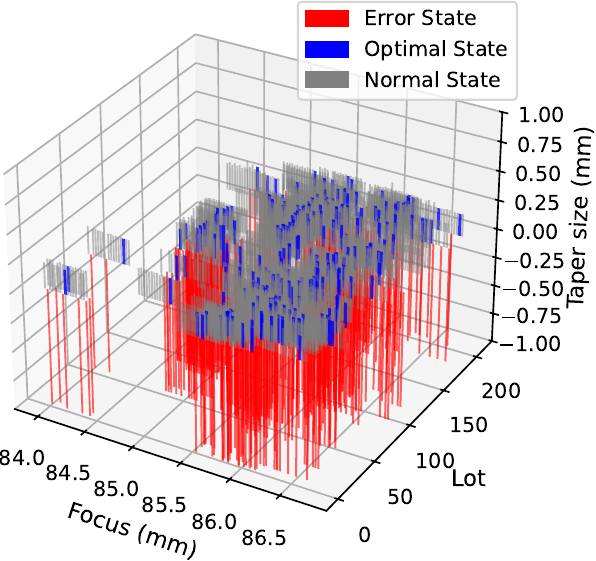}
    \caption{3D parameter space visualization for Stage 3 - Taper Optimization.}
    \label{stage3_param_visual}
    \end{figure}
    
    \item \textbf{Reward Function}: The reward function for Stage 3 is defined as:

    \begin{equation}
    \label{stage3_reward_function}
        R3=
        \begin{cases} 
        -200, \textit{if Error}, \\
        \left(\frac{100}{\textit{taper}}\right) - (\textit{step} \times 0.1), \textit{if Optimal}, \\
        -\left(\frac{1}{\textit{taper}}\right) - (\textit{step} \times 0.1), \textit{if Normal}.
        \end{cases}
    \end{equation}
    Eq. \ref{stage3_reward_function} is designed to minimize taper size in Stage 3, where the agent performs a focused search around the optimal focus value previously found in Stage 1. In the example shown in Fig.~\ref{fig:nps_architecture}, the taper values range from 0.204 mm (\textit{optimal states}) to 0.218 mm (\textit{normal states}), resulting in a maximum theoretical reward of \( \frac{100}{0.204} \approx 490 \) for the best-performing configuration. In contrast, \textit{normal states} result in small negative rewards; for example, taper = 0.214 yields \( -\frac{1}{0.214} \approx -4.67 \). \textit{Error states} (e.g., taper = -1) are assigned a fixed penalty of -200. However, in other lots where the optimal focus range is different, the taper values and the reward bounding may shift accordingly.

\end{itemize}

\section{Experimental results}
\label{section:5}
\subsection{Experimental Setup}
This subsection explains the experimental setup for our evaluation. We have a total of 210 lots, with 90 for training and 120 for testing. The agent is trained on each material sequentially, starting with Material A, followed by A1, and then B, using the $\varepsilon$-greedy policy across these lots. For each lot, the agent runs 100 episodes. In each episode, the agent is allowed a maximum of 100 steps to find the shortest path to the optimal state. During the training phase, the $\varepsilon$ value starts at 1.0 and gradually decreases over the training lots, reaching 0.1. This means that the $\varepsilon$ value remains constant within each lot and only reduces when moving to the next lot, encouraging more exploration in the early lots and more exploitation of learned knowledge in later lots.
Once the training is complete, the agent is evaluated on the remaining 120 testing lots with a fixed $\varepsilon$ of 0.1, ensuring that the agent primarily relies on the learned policies while allowing limited exploration. This setup provides a comprehensive evaluation of the agent's performance across different materials, ensuring effectiveness for optimizing focus, power, and taper settings.

\begin{figure*}[!h]
    \centering
    \begin{subfigure}{0.69\columnwidth}
        \includegraphics[width=\linewidth]{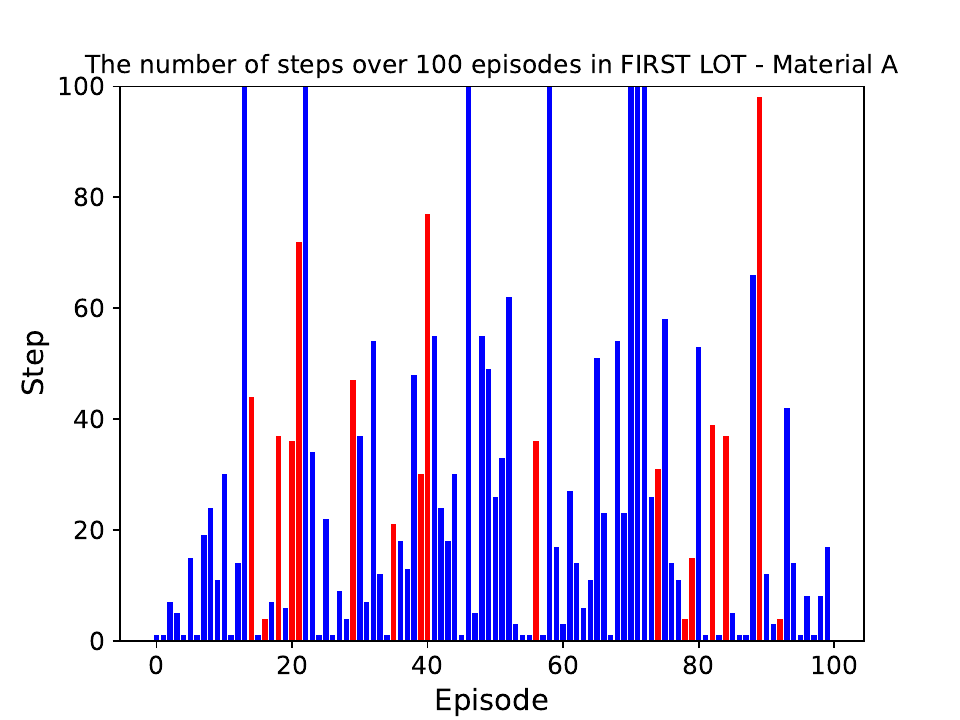}
        \caption{}
        \label{first_a}
    \end{subfigure}
    \begin{subfigure}{0.69\columnwidth}
        \includegraphics[width=\linewidth]{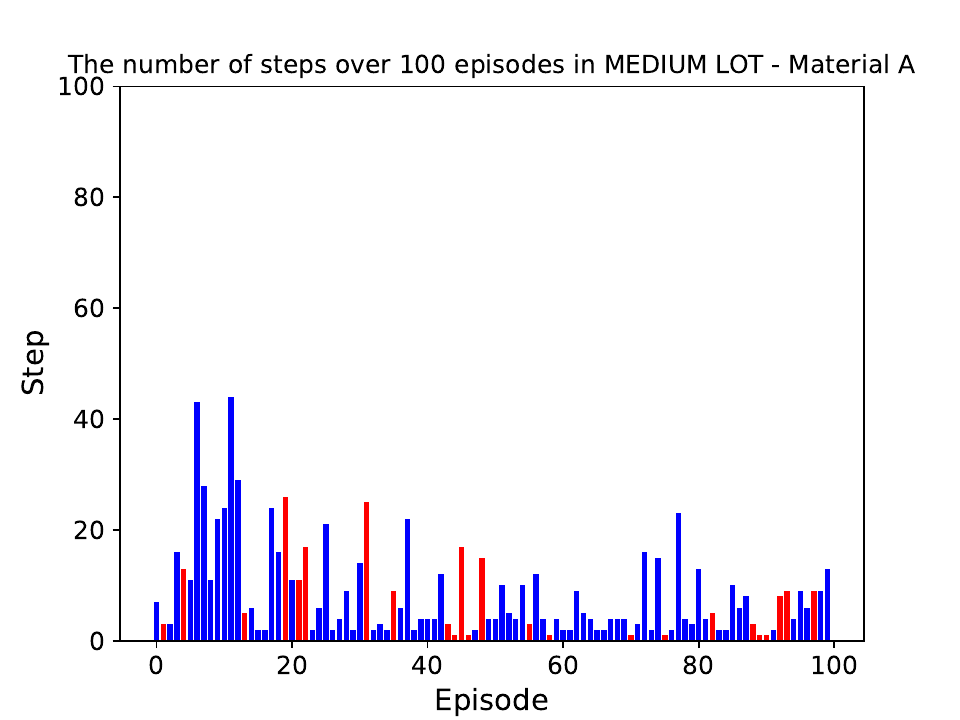}
        \caption{}
        \label{medium_a}
    \end{subfigure}
    \begin{subfigure}{0.69\columnwidth}
        \includegraphics[width=\linewidth]{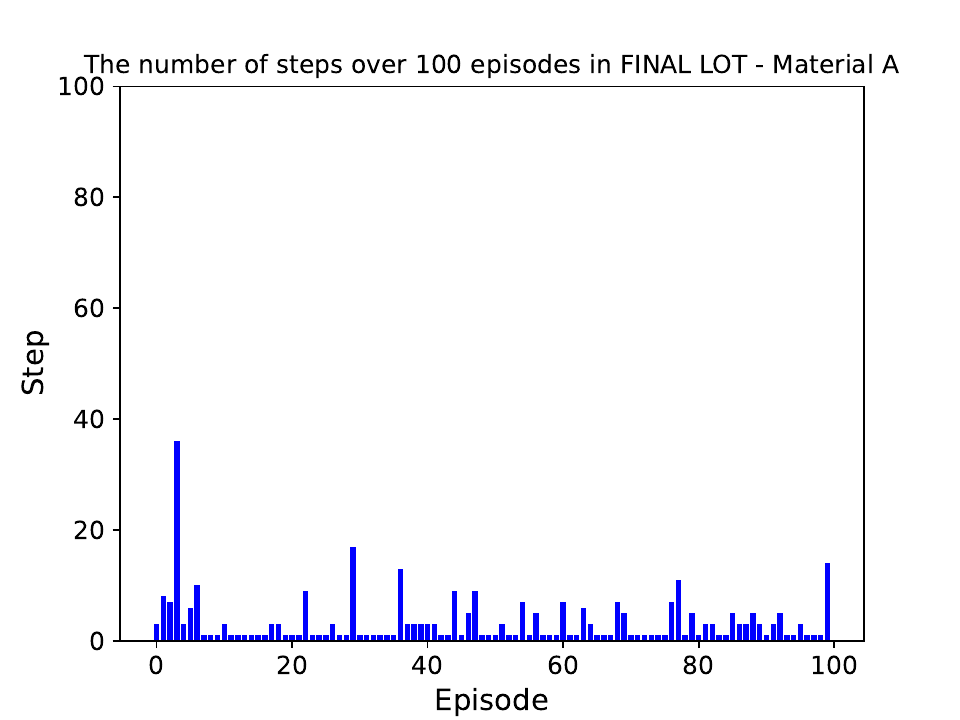}
        \caption{}
        \label{final_a}
    \end{subfigure}   
    
    \begin{subfigure}{0.69\columnwidth}
        \includegraphics[width=\linewidth]{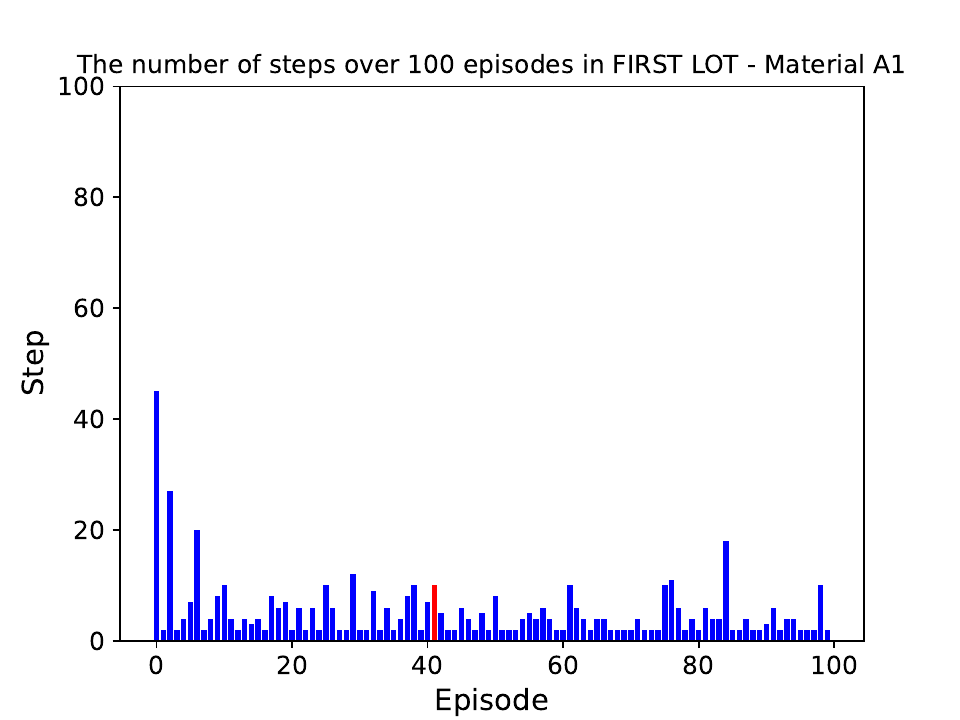}
        \caption{}
        \label{first_a1}
    \end{subfigure}
    \begin{subfigure}{0.69\columnwidth}
        \includegraphics[width=\linewidth]{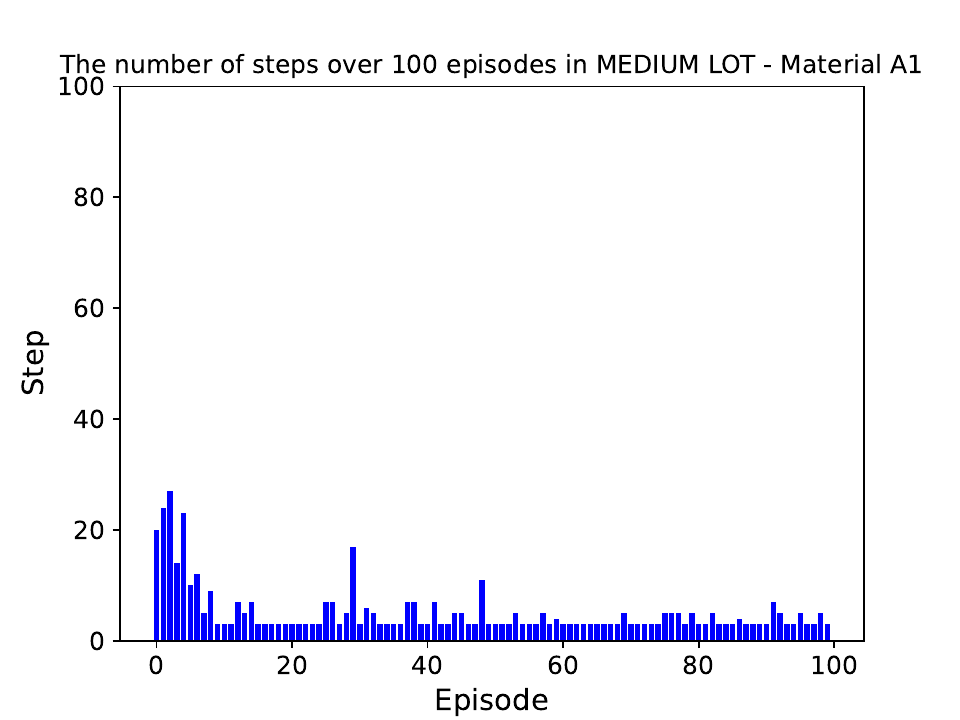}
        \caption{}
        \label{medium_a1}
    \end{subfigure}
    \begin{subfigure}{0.69\columnwidth}
        \includegraphics[width=\linewidth]{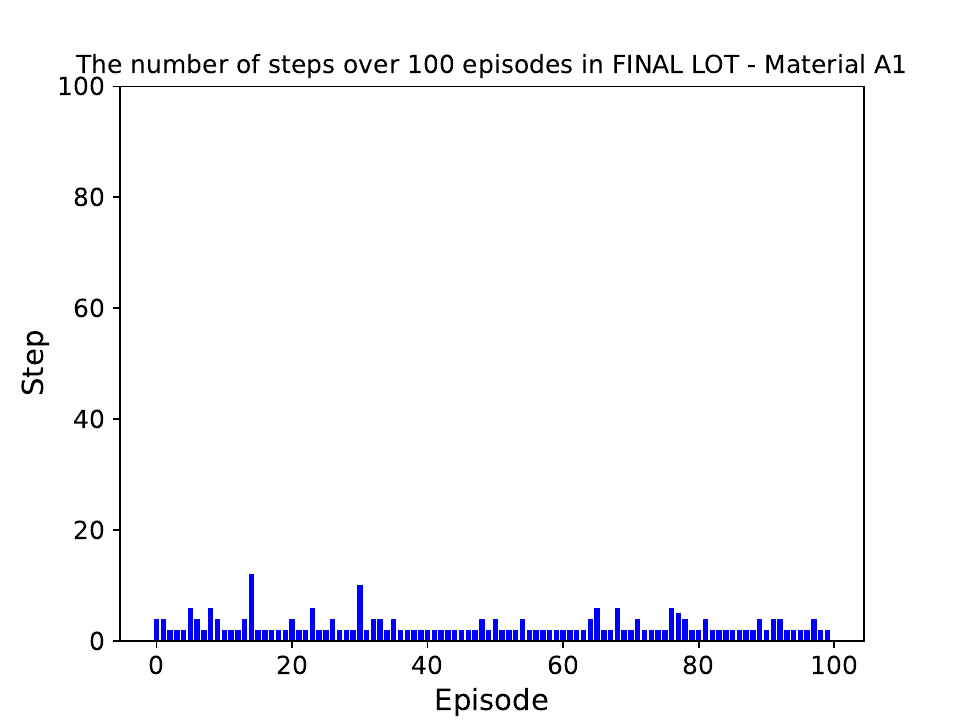}
        \caption{}
        \label{final_a1}
    \end{subfigure}   

    \begin{subfigure}{0.69\columnwidth}
        \includegraphics[width=\linewidth]{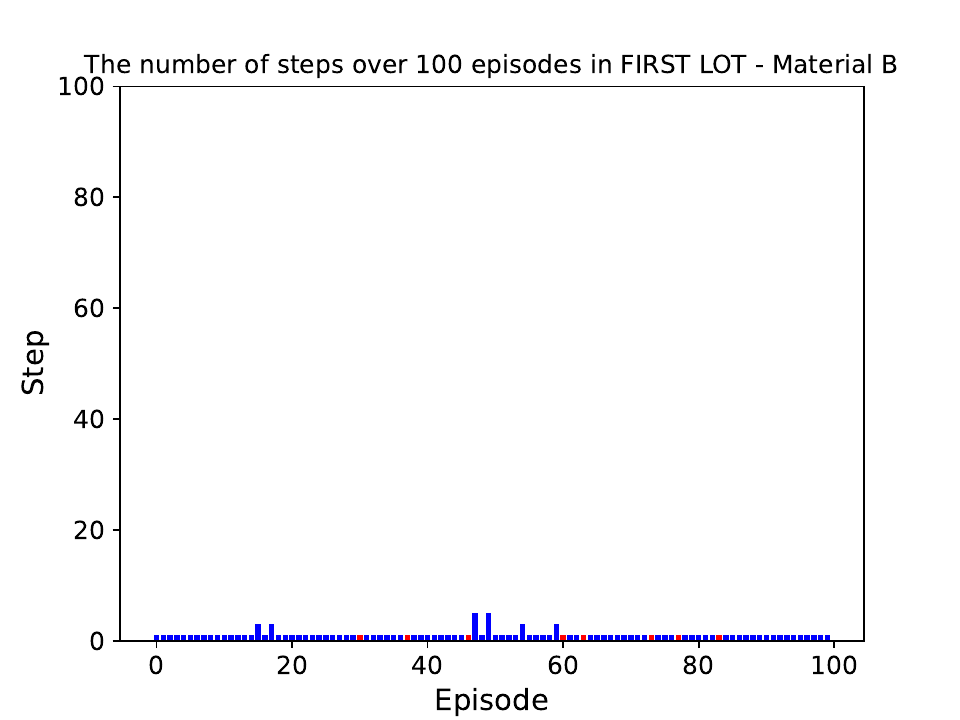}
        \caption{}
        \label{first_b}
    \end{subfigure}
    \begin{subfigure}{0.69\columnwidth}
        \includegraphics[width=\linewidth]{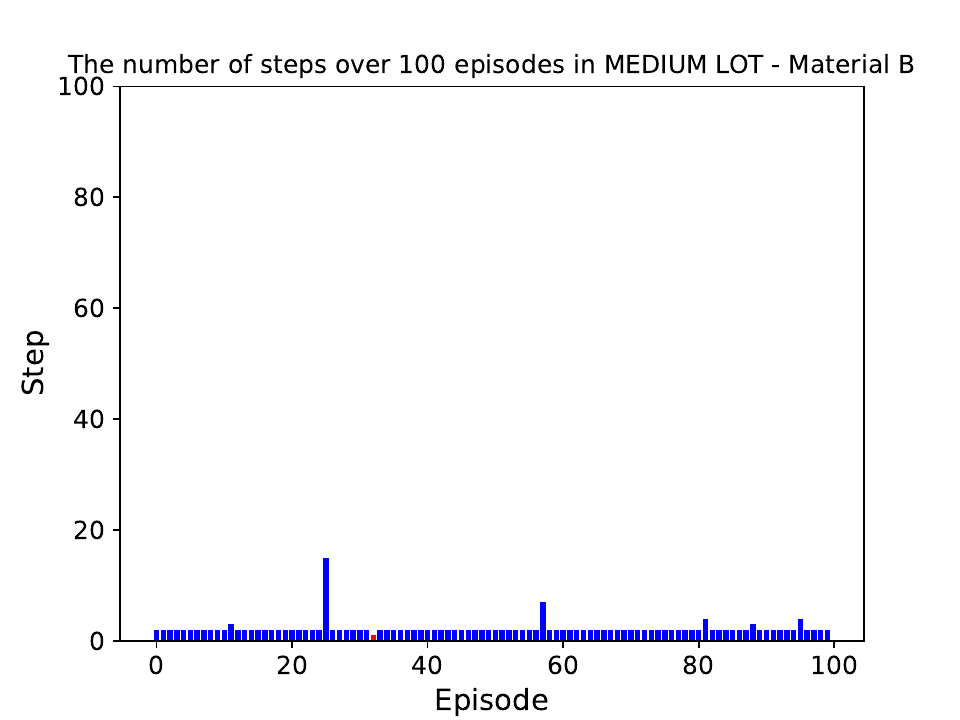}
        \caption{}
        \label{medium_b}
    \end{subfigure}
    \begin{subfigure}{0.69\columnwidth}
        \includegraphics[width=\linewidth]{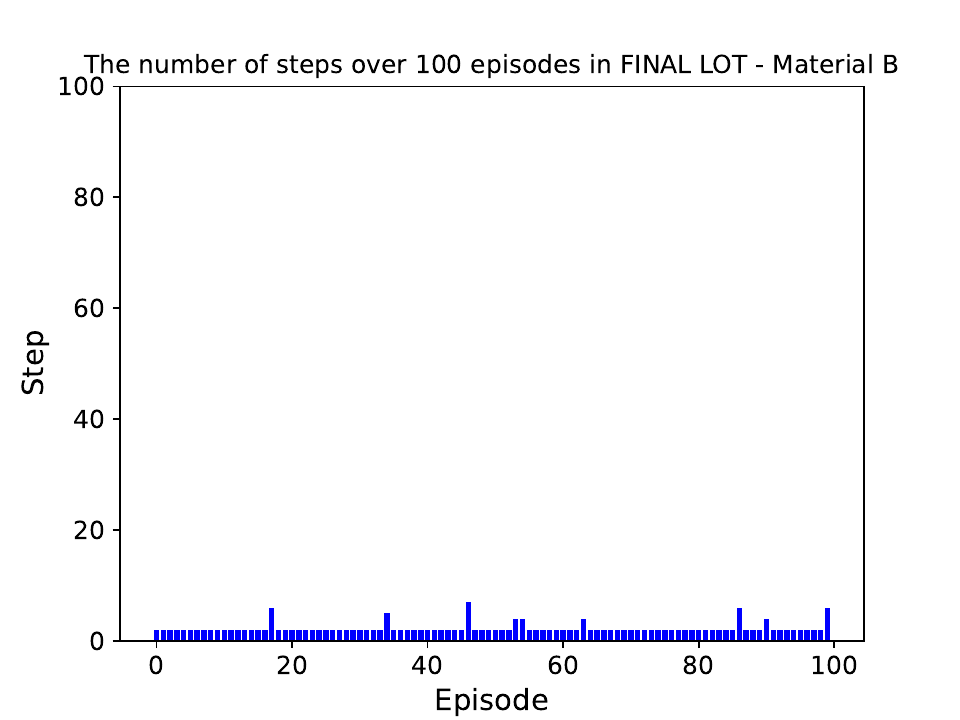}
        \caption{}
        \label{final_b}
    \end{subfigure} 
    \caption{Steps in first(a), medium(b), and final(c) lot in material A; Steps in first(d), medium(e), and final(f) lot in material A1; and Steps in first(g), medium(h), and final(i) lot in material B.}
    \label{figure:3}
\end{figure*}

\subsection{Baselines}
To evaluate the performance of the proposed RL$^{2}$C algorithm, we compare it with three baseline methods: \textbf{Random Search}, \textbf{RL-based Bayesian Optimization} and \textbf{RL-based PSO}. These methods provide straightforward and probabilistic approaches for optimization, serving as benchmarks for assessing the effectiveness of our proposed method.

\textbf{Random Search:} Random Search is a stochastic optimization technique in which the agent selects actions uniformly at random from the defined action space without employing any structured strategy or policy. At each decision step, the agent randomly selects actions in a purely probabilistic manner, with no consideration of previous experiences or rewards, providing a straightforward baseline method. 

\textbf{RL-based Bayesian Optimization} \citep{chang2024optimization}: RL-based Bayesian Optimization integrates RL principles with Bayesian search strategies, leveraging a probabilistic model to guide the action selection process. The agent considers only a few actions to take according to their experience and then chooses the action that would give the maximum expected cumulative reward using the Gaussian process. 

\textbf{RL-based PSO} \citep{zhang2023q}:  Inspired by social behaviors in nature like birds flocking or fish schooling, PSO consists of a swarm of candidate solutions, or particles, in the action space. The agent’s actions are adapted according to its own experiences and the results of other particles in the neighborhood to optimize decision-making.

\subsection{Main Results}

\subsubsection{Step Efficiency Analysis Over Specific Lots}
Fig. \ref{figure:3} illustrates the number of steps required to find the optimal state of the proposed method across different lot positions (first, medium, and final) during the training phase for three distinct materials (A, A1, and B). The figures show the number of steps to reach the optimal state (blue bars) and the number of steps to reach the error state (red bars) over 100 episodes in each lot.

For material A in Fig. \ref{first_a}, the result presents a very high number of steps across episodes. For instance, there are some episodes that take up to 100 steps to find the optimal state, and there are also episodes where the agent reaches an error state during the learning process. This suggests that the agent needs to perform many possible actions in the exploration phase to gain a better understanding of the environment. During this period, the agent may not be able to find the optimal state or may encounter error states. However, after obtaining sufficient knowledge about the current environment, the number of steps has been quickly reduced to under 40 steps in Fig. \ref{medium_a} and under 20 steps in Fig. \ref{final_a}.
This reduction in the number of steps indicates that the agent has successfully learned the optimal policies and can exploit them more efficiently, resulting in a faster convergence towards the optimal state.

For material A1 in Fig. \ref{first_a1}, the agent exhibits a more consistent and stable convergence rate, with fewer spikes and lower variability in the number of steps across episodes. This behavior is particularly evident in the medium and final lots, as shown in Figs. \ref{medium_a1} and \ref{final_a1}. Similarly, for material B in Figs. \ref{first_b}, \ref{medium_b}, and \ref{final_b}, the figures demonstrate an immediate convergence rate, with the agent requiring a very low number of steps to reach the optimal state across episodes. This trend is consistent across all three lots, implying that the agent has successfully learned the optimal policies and can exploit them efficiently for most of the lots in Material B.

\subsubsection{Step Efficiency Analysis Across Different Stages and Materials}
\begin{figure}[!h]
    \centering
    \begin{subfigure}{\columnwidth}
        \includegraphics[width=\textwidth]{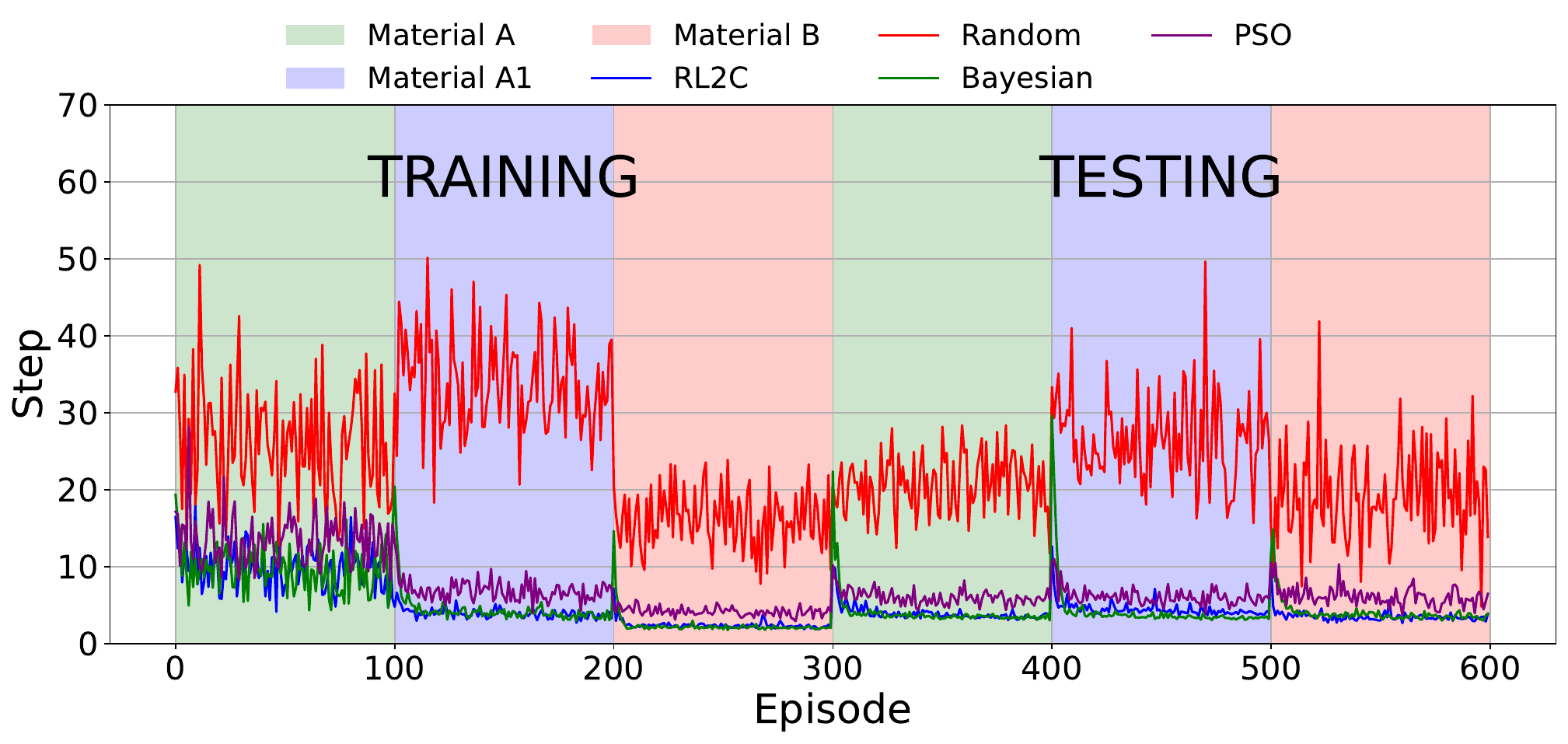}
        \caption{Stage 1 - Focus Optimization}
        \label{fig:step-stage1}
    \end{subfigure}
    \begin{subfigure}{\columnwidth}
        \includegraphics[width=\textwidth]{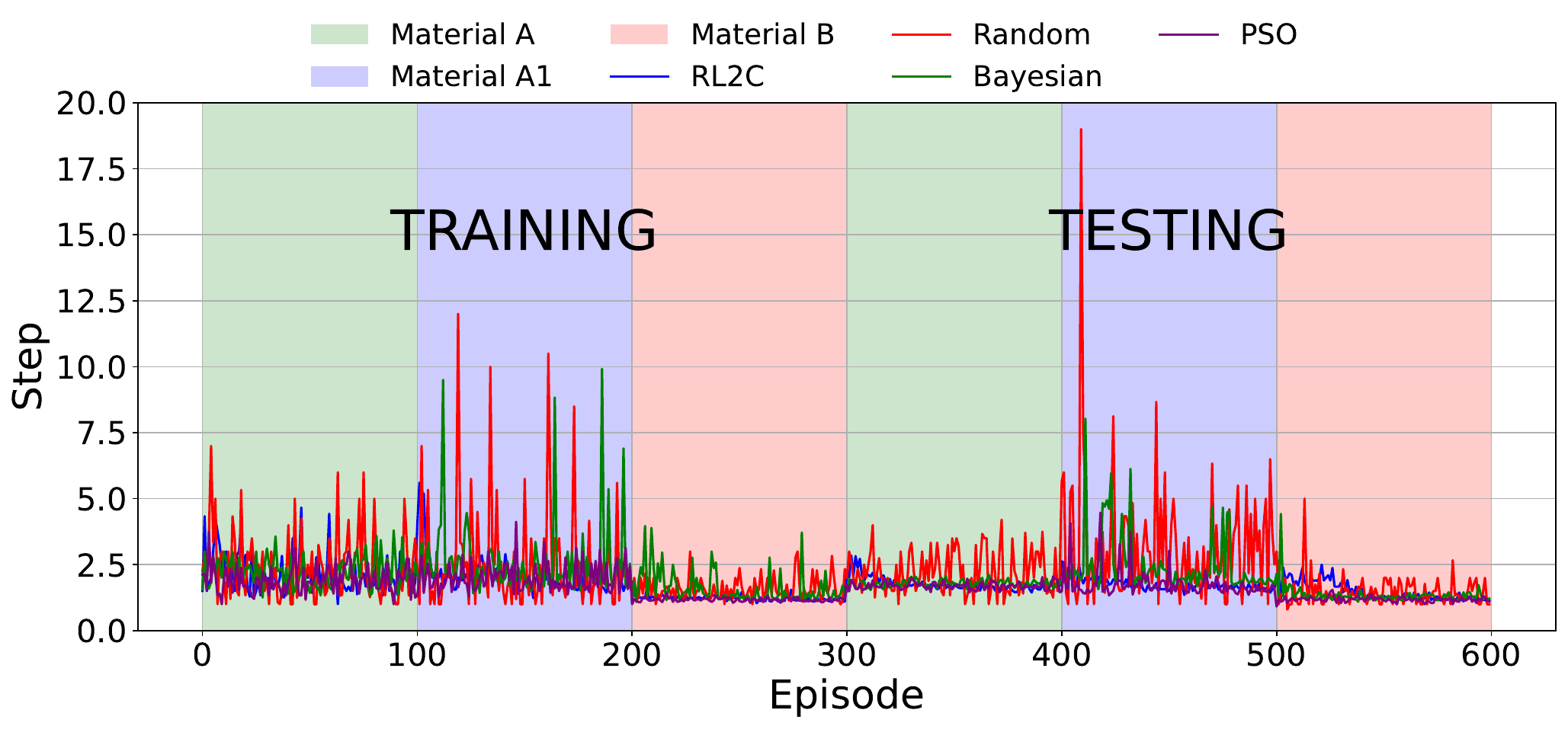}
        \caption{Stage 2 - Power Optimization}
        \label{fig:step-stage2}
    \end{subfigure}
        \begin{subfigure}{\columnwidth}
        \includegraphics[width=\textwidth]{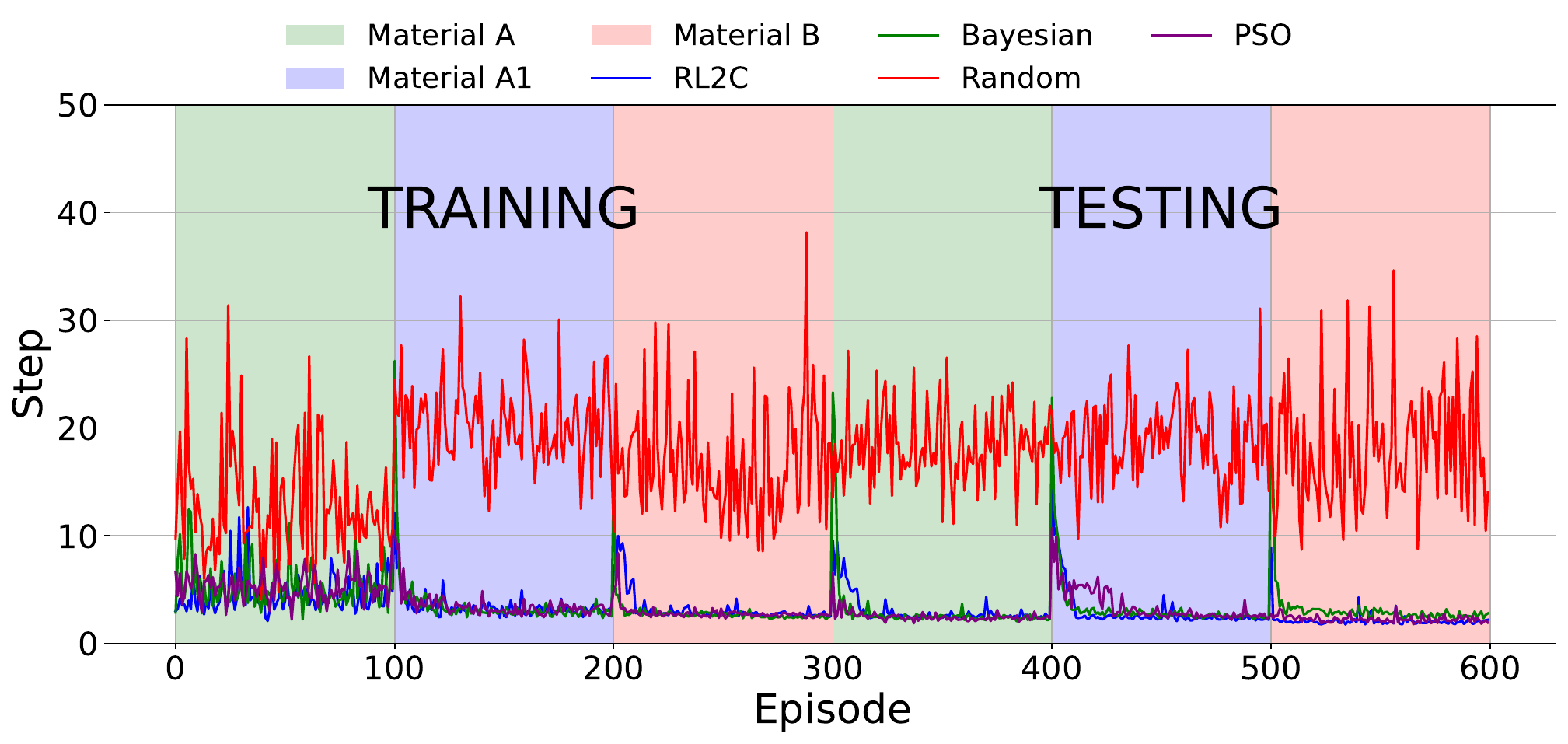}
        \caption{Stage 3 - Taper Optimization}
        \label{fig:step-stage3}
    \end{subfigure}

    \caption{Step comparison of RL$^{2}$C and baseline methods across three optimization stages.}
    \label{fig:step_efficiency}
\end{figure}

Fig.~\ref{fig:step_efficiency} illustrates the average number of steps required by various optimization methods to reach the optimal state across three stages: Focus Optimization, Power Optimization, and Taper Optimization. The results are reported for three different materials (Material A, Material A1, and Material B) during both the training and testing phases. The comparison includes the proposed RL$^{2}$C method and three baseline methods: Random Search, RL-based Bayesian Optimization, and RL-based PSO.

As shown in Fig.~\ref{fig:step-stage1}, during the focus optimization stage, the RL$^{2}$C method consistently requires fewer steps to reach the optimal state compared to the baseline methods. For all materials tested, RL$^{2}$C demonstrates rapid convergence, particularly during the training phase, where it stabilizes around 5 steps by the midpoint of the training episodes. In contrast, Random Search exhibits significant variability throughout both the training and testing phases, often requiring more than 15 steps to achieve optimization. Bayesian Optimization and PSO perform moderately, with PSO showing slightly more consistency than Bayesian Optimization, but both remain less efficient than RL$^{2}$C. 
A noticeable pattern across all methods is the sudden increase in the number of steps required when the agent transitions to a new material (e.g., from Material A to A1 or from A1 to B). This spike occurs at the start of each new material phase, indicating the challenge of adapting to previously unseen conditions. However, RL$^{2}$C quickly recovers from this situation, showcasing great adaptability compared to the baseline methods, which exhibit slower recovery and higher variability.

In Fig.~\ref{fig:step-stage2}, the step efficiency for the power optimization stage shows a similar trend. RL$^{2}$C outperforms the baseline methods across all materials in both the training and testing phases. During training, RL$^{2}$C converges to the optimal state within approximately 3 to 5 steps on average for all three materials, indicating a high level of efficiency in finding the parameter space. Conversely, Random Search requires significantly more steps, fluctuating between 10 and 20 steps throughout the episodes. Bayesian Optimization and PSO demonstrate better performance compared to Random Search but still require more steps on average compared to RL$^{2}$C. 
The results for the taper optimization stage are presented in Fig.~\ref{fig:step-stage3}. Similar to the previous stages, RL$^{2}$C consistently outperforms the baseline methods in terms of step efficiency. During training, RL$^{2}$C reaches the optimal state within 2 to 4 steps on average, with minimal variability across different episodes. 

Across all three optimization stages, RL$^{2}$C demonstrates a clear advantage in terms of step efficiency, particularly in adapting to new materials. The sudden increases in step count observed during material transitions are a critical challenge in real-world industrial processes. The ability of RL$^{2}$C to quickly adapt and stabilize after these transitions highlights its robustness and practical applicability in dynamic environments, where material properties may vary significantly.

\subsubsection{Performance of RL$^{2}$C in Step and Time Efficiency Over Baseline Methods}
Table \ref{tab:rl2c_improvements} presents a performance comparison of the RL$^{2}$C algorithm with baseline methods (RL-Bayesian, RL-PSO, and Random Search) in terms of step and time efficiency across three stages of optimization (Stage 1, Stage 2, and Stage 3). Both training and testing phases are evaluated, and the results highlight the significant advantages of RL$^{2}$C.

For step efficiency, during the training phase, RL$^{2}$C consistently achieves the lowest average number of steps across all three stages. In Stage 1, RL$^{2}$C requires an average of 5.1 steps compared to 5.4 for RL-Bayesian and 25.3 for Random Search. Similarly, in Stage 2 and Stage 3, RL$^{2}$C maintains superior step efficiency with averages of 1.6 and 3.8 steps, respectively. The algorithm demonstrates strong performance in the testing phase as well, with lower average steps in all stages, further confirming its effectiveness in converging to optimal states with minimal exploration.

For time efficiency, RL$^{2}$C shows a significant improvement over the baseline methods in both training and testing phases. In the training phase, RL$^{2}$C completes Stage 1 with an average time of 0.03 seconds, which is significantly faster than RL-Bayesian (4.77 seconds) and RL-PSO (0.11 seconds). Similar trends are observed in Stages 2 and 3, where RL$^{2}$C's average times for both stages are 0.02 seconds. Notably, RL-Bayesian exhibits the highest time cost, particularly in Stage 2, where it requires an average of 11.23 seconds. In the testing phase, RL$^{2}$C maintains superior time efficiency, completing all stages in approximately 0.02 seconds on average, overcoming RL-Bayesian and RL-PSO.

\begin{table*}[h]
    \centering
    \caption{Performance of RL$^{2}$C Compared to baseline methods in step and time efficiency (in seconds).}
    \label{tab:rl2c_improvements}
    \begin{tabular}{|l|c|c|c|c|c|c|c|c|c|}
    \hline
    \textbf{STEP EFFICIENCY} & \multicolumn{3}{c|}{Stage 1} & \multicolumn{3}{c|}{Stage 2} & \multicolumn{3}{c|}{Stage 3} \\ 
    \hline
    \textbf{Training ($\varepsilon$=1$\rightarrow$0.1)} & Avg & Max & Min & Avg & Max & Min & Avg & Max & Min \\ 
    \hline
    \textbf{RL$^{2}$C} & \textbf{5.1} & \textbf{17.2} & \textbf{1.8} & \textbf{1.6} & \textbf{5.6} & \textbf{1.0} & \textbf{3.8} & \textbf{12.7} & \textbf{2.3} \\
    RL-Bayesian & 5.4 & 18.0 & 1.9 & 2.2 & 9.9 & 1.0 & 3.8 & 26.2 & 2.1 \\
    RL-PSO & 8.2 & 28.1 & 2.9 & 1.7 & 4.1 & 1.0 & 3.9 & 10.5 & 2.2 \\
    Random Search & 25.3 & 50.2 & 7.8 & 7.2 & 12.0 & 1.0 & 16.6 & 38.2 & 4.1 \\
    % \textbf{Improvement(\%)} & \textbf{79.8} & \textbf{65.7} & \textbf{76.9} & \textbf{77.8} & \textbf{53.3} & \textbf{0.0} & \textbf{77.1} & \textbf{66.8} & \textbf{43.9} \\ 

    \hline
    \textbf{Testing ($\varepsilon$=0.1)} & Avg & Max & Min & Avg & Max & Min & Avg & Max & Min \\ 
    \hline
    \textbf{RL$^{2}$C} & \textbf{4.0} & \textbf{12.6} & \textbf{2.7} & \textbf{1.7} & \textbf{2.8} & \textbf{1.1} & \textbf{2.8} & \textbf{15.3} & \textbf{1.8} \\
    RL-Bayesian & 4.3 & 29.6 & 3.0 & 1.9 & 8.0 & 1.1 & 3.2 & 23.3 & 2.1 \\
    RL-PSO & 6.1 & 19.9 & 3.2 & 1.5 & 4.5 & 1.0 & 4.4 & 10.7 & 1.8 \\
    Random Search & 22.2 & 49.7 & 4.8 & 11.3 & 19.0 & 1.2 & 18.4 & 34.6 & 8.7 \\
    % \textbf{Improvement(\%)} & \textbf{81.9} & \textbf{74.7} & \textbf{43.8} & \textbf{84.9} & \textbf{85.3} & \textbf{8.3} & \textbf{84.7} & \textbf{55.7} & \textbf{79.3} \\

    %%%%%%%%%%%%%%%%%%%%%%%%%%%%%%%%%%%%%%%%%%%%%%
    \hline
    \textbf{TIME EFFICIENCY} & \multicolumn{3}{c|}{Stage 1} & \multicolumn{3}{c|}{Stage 2} & \multicolumn{3}{c|}{Stage 3} \\ 
    \hline
    \textbf{Training ($\varepsilon$=1$\rightarrow$0.1)} & Avg & Max & Min & Avg & Max & Min & Avg & Max & Min \\ 
    \hline
    \textbf{RL$^{2}$C} & \textbf{0.03} & \textbf{0.10} & \textbf{0.01} & \textbf{0.02} & \textbf{0.09} & \textbf{0.01} & \textbf{0.02} & \textbf{0.10} & \textbf{0.01} \\
    RL-Bayesian & 4.77 & 9.41 & 0.10 & 11.23 & 40.36 & 0.02 & 8.44 & 41.85 & 0.01 \\
    RL-PSO & 0.11 & 0.20 & 0.05 & 0.09 & 0.81 & 0.01 & 0.10 & 0.24 & 0.01 \\
    Random Search & 0.08 & 0.16 & 0.02 & 0.02 & 0.08 & 0.01 & 0.05 & 0.20 & 0.01 \\
    % \textbf{Improvement(\%)} & \textbf{99.3} & \textbf{98.9} & \textbf{90.0} & \textbf{99.8} & \textbf{99.7} & \textbf{50.0} & \textbf{99.7} & \textbf{99.7} & \textbf{0.0} \\

    \hline
    \textbf{Testing ($\varepsilon$=0.1)} & Avg & Max & Min & Avg & Max & Min & Avg & Max & Min \\ 
    \hline
    \textbf{RL$^{2}$C} & \textbf{0.02} & \textbf{0.08} & \textbf{0.01} & \textbf{0.02} & \textbf{0.08} & \textbf{0.01} & \textbf{0.02} & \textbf{0.11} & \textbf{0.01} \\
    RL-Bayesian & 5.77 & 19.97 & 0.02 & 13.22 & 37.83 & 0.01 & 10.02 & 50.90 & 0.01 \\
    RL-PSO & 0.11 & 0.29 & 0.06 & 0.13 & 1.41 & 0.01 & 0.11 & 1.09 & 0.01 \\
    Random Search & 0.06 & 0.16 & 0.02 & 0.02 & 0.08 & 0.01 & 0.06 & 0.18 & 0.01 \\
    % \textbf{Improvement(\%)} & \textbf{99.65} & \textbf{99.6} & \textbf{50.0} & \textbf{99.8} & \textbf{99.7} & \textbf{0.0} & \textbf{99.8} & \textbf{99.7} & \textbf{0.0} \\
    \hline
    \end{tabular}
\end{table*}

The result demonstrates that RL$^{2}$C is not only capable of efficiently exploring and exploiting the parameter space but also exhibits superior generalization to new conditions, as evidenced by its consistent performance across both training and testing phases. Its ability to achieve optimal states with fewer steps and significantly lower computational time highlights its suitability for real-time optimization tasks, making it a robust solution for laser cutting parameter optimization. The consistent performance across all stages and phases further emphasizes its adaptability and scalability in diverse scenarios.

\section{Conclusion}
\label{section:6}
This study presents a comprehensive evaluation of the proposed RL$^{2}$C algorithm for optimizing laser cutting parameters across three vital stages: Focus Optimization, Power Optimization, and Taper Optimization. The results demonstrate that RL$^{2}$C consistently outperforms baseline methods, including Random Search, RL-based Bayesian Optimization, and RL-based PSO, in terms of step efficiency, time efficiency, and adaptability to dynamic environments. Across all tested materials (Material A, Material A1, and Material B), RL$^{2}$C exhibits faster convergence, maintaining low execution times and requiring fewer steps to reach the optimal state.

A key strength of RL$^{2}$C lies in its ability to handle transitions between materials efficiently. While baseline methods, particularly Bayesian Optimization, suffer from significant time and step overheads during these transitions, RL$^{2}$C adapts rapidly with minimal computational cost. This adaptability underscores its robustness and practicality for real-world applications, where material properties can vary significantly between production lots.

In summary, RL$^{2}$C emerges as a highly efficient and scalable solution for laser cutting parameter optimization, demonstrating optimal performance in both static and dynamic environments. These findings pave the way for its deployment in industrial settings, where efficiency, consistency, and adaptability are important. Future work will explore extending RL$^{2}$C to additional manufacturing processes and integrating it with other advanced learning frameworks to further enhance its applicability and scalability.

\bmhead{Author contributions}
Quan Khanh Pham conducts the experiments, coding, and manuscript writing. Majid Kundroo assists in the coding and manuscript preparation. Geunwoo Ban, Seongho Bae, and Taehong Kim provide technical guidance, contribute to experimental design, revise the manuscript, and supervise the research.

\bmhead{Funding} 
This work was supported by the Innovative Human Resource Development for Local Intellectualization program through the Institute of Information \& Communications Technology Planning \& Evaluation (IITP) grant funded by the Korea government (MSIT) (IITP-2025-RS-2020-II201462).

\section*{Declarations}

% Some journals require declarations to be submitted in a standardised format. Please check the Instructions for Authors of the journal to which you are submitting to see if you need to complete this section. If yes, your manuscript must contain the following sections under the heading `Declarations':
\bmhead{Conflict of interest}  
The authors declare that they have no conflict of interest.

\bmhead{Data availability}  
The NPS dataset used in this study was acquired from the NPS, and they have not given their permission for researchers to share their data. Data requests can be made to the NPS via this email:  nps@npstech.co.kr

\bibliography{sn-bibliography.bib}% common bib file

\end{document}